\documentclass{article} 
\usepackage{iclr2027_conference,times}

\usepackage{amsmath,amsfonts,bm}

\def\eqref#1{equation~\ref{#1}}

\def\1{\bm{1}}

\DeclareMathAlphabet{\mathsfit}{\encodingdefault}{\sfdefault}{m}{sl}
\SetMathAlphabet{\mathsfit}{bold}{\encodingdefault}{\sfdefault}{bx}{n}

\usepackage{amsmath}
\usepackage{xcolor}
\usepackage{tcolorbox}
\usepackage{adjustbox}
\definecolor{BPgood}{HTML}{176343}
\definecolor{BPbad}{HTML}{A32638}
\definecolor{BPcheck}{HTML}{875A0A}
\newcommand{\BPGood}[1]{\par\smallskip{\color{BPgood}\bfseries #1}\par}
\newcommand{\BPBad}[1]{\par\smallskip{\color{BPbad}\bfseries #1}\par}
\newcommand{\BPCheck}[1]{\par\smallskip{\color{BPcheck}\bfseries #1}\par}
\newcommand{\BPGap}{\par\smallskip{\footnotesize\color{gray}[Intermediate steps omitted.]}\par\smallskip}
\newcommand{\BPCase}[4]{%
\begin{tcolorbox}[title={#1},fonttitle=\bfseries,
 colback=white,colframe=black!25,colbacktitle=black!5,coltitle=black,
 boxrule=0.5pt,arc=1mm,left=2mm,right=2mm,top=2mm,bottom=2mm]
\small\setlength{\parindent}{0pt}
\setlength{\abovedisplayskip}{4pt}\setlength{\belowdisplayskip}{4pt}
\textbf{Problem.} #2\par\medskip\hrule\medskip
\begin{minipage}[t]{0.48\linewidth}\vspace{0pt}\raggedright
\textbf{Base (step 0)}\hfill{\color{BPbad}\bfseries Incorrect}\par #3
\end{minipage}\hfill
\begin{minipage}[t]{0.48\linewidth}\vspace{0pt}\raggedright
\textbf{PreSO (step 500)}\hfill{\color{BPgood}\bfseries Correct}\par #4
\end{minipage}
\par\medskip\hrule\smallskip
\end{tcolorbox}}

\newcommand{\BasePreSOOptimization}{%
\BPCase{Case 1: Preserving the objective in constrained optimization}
{Find the largest possible real part of
$(75+117i)z+(96+144i)/z$ subject to $|z|=4$.
\par
\textbf{Reference answer: $540$.}}
{\BPGood{1. The constraint is represented correctly}
\[
z=x+iy,\qquad x^2+y^2=16.
\]
\BPBad{2. A sign error changes the objective}
The response expands the first term as
\[
\begin{aligned}
(75+117i)(x+iy)
&=75x+\textcolor{BPbad}{117y}\\
&\quad+i(75y+117x).
\end{aligned}
\]
Combining this with the correctly computed second real part gives
the incorrect objective
\[
81x+126y.
\]
\BPBad{3. Optimization is applied to the wrong objective}
\[
4\sqrt{81^2+126^2}\approx\boxed{599.16}.
\]
}
{\BPGood{1. Correct real-part reduction}
The two terms contribute $75x-117y$ and $6x+9y$:
\[
\operatorname{Re}(E)=81x-108y.
\]
\BPGood{2. Parameterization preserves the constraint}
\[
x=4\cos\theta,\qquad y=4\sin\theta.
\]
Thus the objective becomes
\[
f(\theta)=324\cos\theta-432\sin\theta.
\]
\BPGood{3. Compute the extremal value}
Using the amplitude of a linear combination of sine and cosine:
\[
\sqrt{324^2+(-432)^2}=540.
\]
\BPGood{4. Discuss attainability}
The response explains that the angle can be chosen to align
$(\cos\theta,\sin\theta)$ with $(324,-432)$, so the maximum
is achievable.
\[
\boxed{540}.
\]
}
}

\newcommand{\BasePreSOTangency}{%
\BPCase{Case 2: Revising a geometric representation}
{Find all real $k$ for which
$|25+20i-z|=5$ and $|z-4-k|=|z-3i-k|$
have exactly one complex solution $z$.
Write the sum of these $k$ values as $m/n$ in lowest terms and
find $m+n$. 
\par
\textbf{Reference answer: $77$.}}
{\BPGood{1. Correct geometric interpretation}
The first locus is a circle centered at $(25,20)$ with radius 5.
The second is a perpendicular bisector. Exactly one intersection
requires tangency.
\BPBad{2. The real parameter is placed on the wrong axis}
The response represents $A=4+k$ and $B=k+3i$ as
\[
A=(4,k),\qquad B=(0,k+3).
\]
\BPGap
\BPBad{3. The incorrect geometry is carried forward}
After further rearrangement, the response uses
\[
8x-6y-(7+6k)=0,
\]
and imposes
\[
\frac{|73-6k|}{10}=5.
\]
It obtains $k=23/6,\ 41/2$, hence
\[
\sum k=\frac{73}{3},\qquad\boxed{76}.
\]
}
{\BPBad{1. Initial direction of the bisector is incorrect}
The response incorrectly treats the segment as vertical and
its perpendicular bisector as
\[
y=\frac32.
\]
\BPCheck{2. An incompatible distance is detected}
\[
\left|20-\frac32\right|=18.5\ne5.
\]
``So, something is wrong in our reasoning.''
\BPGap
\BPGood{3. Rebuild the equation from squared distances}
\[
\begin{aligned}
&(x-4-k)^2+y^2\\
&\qquad=(x-k)^2+(y-3)^2.
\end{aligned}
\]
This gives $4x-3y=7/2+4k$.
\BPGood{4. Apply tangency to the repaired line}
\[
\frac{|73/2-4k|}{5}=5
\quad\Longrightarrow\quad
k\in\left\{\frac{23}{8},\frac{123}{8}\right\}.
\]
\[
\sum k=\frac{73}{4},\qquad\boxed{77}.
\]
}
}

\newcommand{\BasePreSOCombinedCases}{%
\begin{figure*}[tp]
\centering
\caption{Case study on two problems with Qwen3-1.7B.}
\label{fig:case-study}
\begin{adjustbox}{max totalsize={\textwidth}{0.96\textheight}}
\begin{minipage}{\textwidth}
\BasePreSOOptimization
\par\vspace{2mm}
\BasePreSOTangency
\end{minipage}
\end{adjustbox}
\end{figure*}%
}

\usepackage{amsmath}
\usepackage{xcolor}
\usepackage{tcolorbox}

\newcommand{\PRPromptFigure}[2]{%
\begin{figure*}[h]
\centering
\begin{tcolorbox}[
 title={#1},fonttitle=\bfseries,
 colback=white,colframe=black!25,colbacktitle=black!5,coltitle=black,
 boxrule=0.5pt,arc=1mm,left=2mm,right=2mm,top=2mm,bottom=2mm]
\small
\setlength{\parindent}{0pt}
\setlength{\abovedisplayskip}{4pt}
\setlength{\belowdisplayskip}{4pt}
#2
\end{tcolorbox}
\end{figure*}%
}

\newcommand{\TeacherModelPrompt}{%
\PRPromptFigure
{Prompts for teacher model}
{
Problem: \{problem\}\par
\medskip

Here is the reference solution for this problem:\par
=== Reference Solution Begin ===\par
\{privileged information\}\par
=== Reference Solution End ===\par
\medskip

After reading the reference solution above, make sure you truly
understand the reasoning behind each step - do not copy or
paraphrase it. Now, using your own words and independent reasoning,
derive the same final answer to the problem above. Think step by
step, explore different approaches, and don't be afraid to
backtrack or reconsider if something doesn't work out:\par
\medskip

Please reason step by step, and put your final answer within
\textbackslash boxed\{\}.
}
}

\newcommand{\HintGenerationPrompt}{%
\PRPromptFigure
{Prompt for hint generation}
{

You are a math tutoring assistant. Given a math problem and its
full solution, your task is to generate a \textbf{brief, general hint}
that guides the solver toward the right approach WITHOUT revealing
specific steps, intermediate results, or the final answer.
\par\medskip

\textbf{Requirements}
\par\smallskip

1. \textbf{Abstraction}: The hint should describe the general
mathematical strategy or key insight, not specific calculations.
It should be applicable to a class of similar problems.
\par

2. \textbf{No answer leakage}: Do NOT include any numerical results,
intermediate values, or the final answer.
\par

3. \textbf{No step-by-step details}: Do NOT outline the solution
procedure. Only indicate WHAT approach to consider, not HOW to
execute it.
\par

4. \textbf{Conciseness}: The hint should be 1--3 sentences.
\par

5. \textbf{Knowledge-level guidance}: Focus on which mathematical
concept, theorem, or technique is relevant.
\par\medskip

\textbf{Output Format}
\par\smallskip

\texttt{Hint: <your hint here>}
\par\medskip

\textbf{Examples}
\par\smallskip

\textbf{Example 1}
\par\smallskip

\textbf{Problem}: Find the number of ways to arrange 5 distinct
books on a shelf such that two specific books are always adjacent.
\par\smallskip

\textbf{Full Solution}: Treat the two specific books as a single
unit, giving 4 units to arrange (\(4! = 24\) ways). The two books
within the unit can be arranged in \(2! = 2\) ways.
Total \(= 24 \times 2 = 48\).
\par\smallskip

\textbf{Generated Hint}:
\par
\texttt{Hint:} Consider treating the constrained elements as a
single unit to reduce the problem, then account for the internal
arrangements within that unit.
\par\medskip

\textbf{Example 2}
\par\smallskip

\textbf{Problem}: Evaluate the integral
\(\int_{0}^{\pi} \sin^{3}(x)\,dx\).
\par\smallskip

\textbf{Full Solution}: Use the identity
\(\sin^{3}(x) = \sin(x)(1 - \cos^{2}(x))\),
then substitute \(u = \cos(x)\), \(du = -\sin(x)\,dx\).
The integral becomes
\(\int_{1}^{-1} -(1-u^{2})\,du\),
which equals
\(\int_{-1}^{1} (1-u^{2})\,du\).
Evaluating gives
\(\bigl[u-u^{3}/3\bigr]_{-1}^{1} = 4/3\).
\par\smallskip

\textbf{Generated Hint}:
\par
\texttt{Hint:} Try reducing the power of the trigonometric function
using a Pythagorean identity, which will set up a straightforward
substitution.
\par\medskip

\textbf{Task}
\par\smallskip

\textbf{Problem}: \{\{problem\}\}
\par
\textbf{Full Solution}: \{\{solution\}\}
\par\medskip

Generate a hint following the requirements above.

}
}

\newcommand{\ConceptGenerationPrompt}{%
\PRPromptFigure
{Prompt for concept generation}
{

You are a math education assistant. Given a math problem and its
full solution, your task is to identify the
\textbf{underlying mathematical concept or topic}
required to solve the problem.
\par\medskip

\textbf{Requirements}
\par\smallskip

1. \textbf{Abstraction}: The concept should describe the broad
mathematical area or topic involved, not the specific solving
method, reasoning process, or solution strategy.
\par

2. \textbf{No answer leakage}: Do NOT include any numerical results,
intermediate calculations, formulas derived from the solution,
or the final answer.
\par

3. \textbf{No solution guidance}: Do NOT provide hints, strategies,
or step-by-step reasoning. Only identify the relevant mathematical
category.
\par

4. \textbf{Generality}: The concept should be applicable to a class
of similar mathematical problems.
\par

5. \textbf{Conciseness}: The output should be a short phrase or
one sentence.
\par\medskip

\textbf{Output Format}
\par\smallskip

\texttt{Concept: \textless the underlying mathematical concept\textgreater}
\par\medskip

\textbf{Example 1}
\par\smallskip

\textbf{Problem}: Find the number of ways to arrange 5 distinct
books on a shelf such that two specific books are always adjacent.
\par\smallskip

\textbf{Full Solution}: Treat the two specific books as a single
unit, giving 4 units to arrange (\(4! = 24\) ways). The two books
within the unit can be arranged in \(2! = 2\) ways.
Total \(= 24 \times 2 = 48\).
\par\smallskip

\textbf{Generated Concept}:
\par
\texttt{Concept:} Permutations with constraints.
\par\medskip

\textbf{Example 2}
\par\smallskip

\textbf{Problem}: Evaluate the integral
\(\int_{0}^{\pi} \sin^{3}(x)\,dx\).
\par\smallskip

\textbf{Full Solution}: Use the identity
\(\sin^{3}(x) = \sin(x)(1 - \cos^{2}(x))\),
then substitute \(u = \cos(x)\), \(du = -\sin(x)\,dx\).
The integral becomes
\(\int_{1}^{-1} -(1-u^{2})\,du\),
which equals
\(\int_{-1}^{1} (1-u^{2})\,du\).
Evaluating gives
\(\bigl[u-u^{3}/3\bigr]_{-1}^{1} = 4/3\).
\par\smallskip

\textbf{Generated Concept}:
\par
\texttt{Concept:} Integration of trigonometric functions.
\par\medskip

\textbf{Task}
\par\smallskip

\textbf{Problem}: \{\{problem\}\}
\par\smallskip

\textbf{Full Solution}: \{\{solution\}\}
\par\medskip

Generate a concept following the requirements above.

}
}

\usepackage{hyperref}
\usepackage{placeins}
\usepackage{url}
\usepackage{graphicx}
\usepackage{wrapfig}
\usepackage{amssymb}
\usepackage{amsthm}
\usepackage{tcolorbox}
\usepackage{enumitem}
\tcbuselibrary{most}

\usepackage{algorithm}
\usepackage{algpseudocode}
\usepackage{multirow}
\usepackage{booktabs}
\usepackage{subcaption}
\usepackage[table]{xcolor}

\usepackage{tabularx}
\usepackage{array}
\newcolumntype{Y}{>{\centering\arraybackslash}X}

\definecolor{lightblue}{RGB}{220,235,250}
\definecolor{mygreen}{RGB}{35,120,45}
\definecolor{myred}{RGB}{180,30,40}
\definecolor{lightgraytext}{RGB}{170,170,170}
\newcommand{\uprise}[1]{\textcolor{mygreen}{\small\;{$\uparrow$ #1}}}
\newcommand{\down}[1]{\textcolor{myred}{\small\;{$\downarrow$ #1}}}
\newcommand{\basex}[1]{\textcolor{lightgraytext}{\small\;{$\uparrow$ #1}}}
\usepackage{tcolorbox}
\tcbuselibrary{skins,breakable}
\newtcolorbox{promptbox}[1]{
  enhanced,
  breakable,
  width=\linewidth,
  title={#1},
  colback=gray!3,
  colframe=gray!50,
  colbacktitle=gray!50,
  coltitle=white,
  fonttitle=\normalsize\bfseries,
  fontupper=\small,
  boxrule=0.5pt,
  titlerule=0pt,
  arc=2mm,
  left=3mm,
  right=3mm,
  top=2mm,
  bottom=2mm,
  before upper={
    \setlength{\parindent}{0pt}
    \setlength{\parskip}{2pt}
    \raggedright
  }
}

\newcommand{\ours}{PreSO}
\newcommand{\oursfull}{Privileged Representation Isolation for On-Policy Self-Distillation}
\title{\ours{}: \oursfull{}}

\author{
\textbf{Yuhang Yang\textsuperscript{1,2}}\thanks{ This work is done during Yuhang Yang’s internship at ByteDance. \texttt{yangyuhang@zju.edu.cn}}\quad 
\textbf{Mingxuan Xia\textsuperscript{1}}\quad 
\textbf{Chao Ye\textsuperscript{2}}\quad 
\textbf{Shuai Zhu\textsuperscript{2}}\quad
\textbf{Shouqing Yang\textsuperscript{1}}
\\[0.3em]
\textbf{Haobo Wang\textsuperscript{1}}\thanks{Corresponding author. \texttt{wanghaobo@zju.edu.cn}, \texttt{wangsiqing.jacky@bytedance.com}}\quad 
\textbf{Shenzhi Yang\textsuperscript{1}}\quad 
\textbf{Zhengqing Zang\textsuperscript{1}}\quad 
\textbf{Chenglong Wang\textsuperscript{2}}\footnotemark[2]
\\[0.4em]
\textsuperscript{1}Zhejiang University\quad
\textsuperscript{2}ByteDance \\
}

\iclrfinalcopy

\begin{document}

\maketitle

\begin{abstract}

On-policy self-distillation (OPSD) uses a self-teacher conditioned on
privileged information to turn the sparse outcome signal of reinforcement
learning with verifiable rewards (RLVR) into dense token-level supervision.
Yet the teacher's advantage may depend on information unavailable to the
student at inference, raising a central question: \textit{which privileged
changes are worth distilling?} We use cross-problem recurrence as an
empirical cue for potentially reusable guidance. Comparing the
representation shifts induced by full solutions and compact hints, we find
that hint-induced shifts are more concentrated and directionally consistent
across problems, and yield more sustained downstream gains despite
revealing less information. Motivated by this association, we propose
\textbf{\oursfull{} (\ours{})}. At the source, \ours{} uses compact hints
to limit the solution-specific detail entering the privileged shift. At
the target, it estimates a leading low-rank subspace from shifts pooled
across related problems, projects the privilege-induced correction onto
this subspace, and adds it to the student's unprivileged representation
to form the representation-level distillation target. The selected
subspace serves as an operational proxy for recurring structure, rather
than a guaranteed separation of transferable and instance-specific
information. On five competition-level mathematics benchmarks, \ours{}
consistently outperforms strong baselines across Qwen3-1.7B, 4B, and 8B,
improving average accuracy over the base model by $\mathbf{6.8}$,
$\mathbf{9.0}$, and $\mathbf{10.0}$ points. Beyond peak accuracy, \ours{}
attains the highest final-checkpoint average at every model scale, with
improvements that persist through the end of training.
\end{abstract}
\vspace{-5pt}

\begin{figure*}[htbp]
\centering
\includegraphics[width=\linewidth]{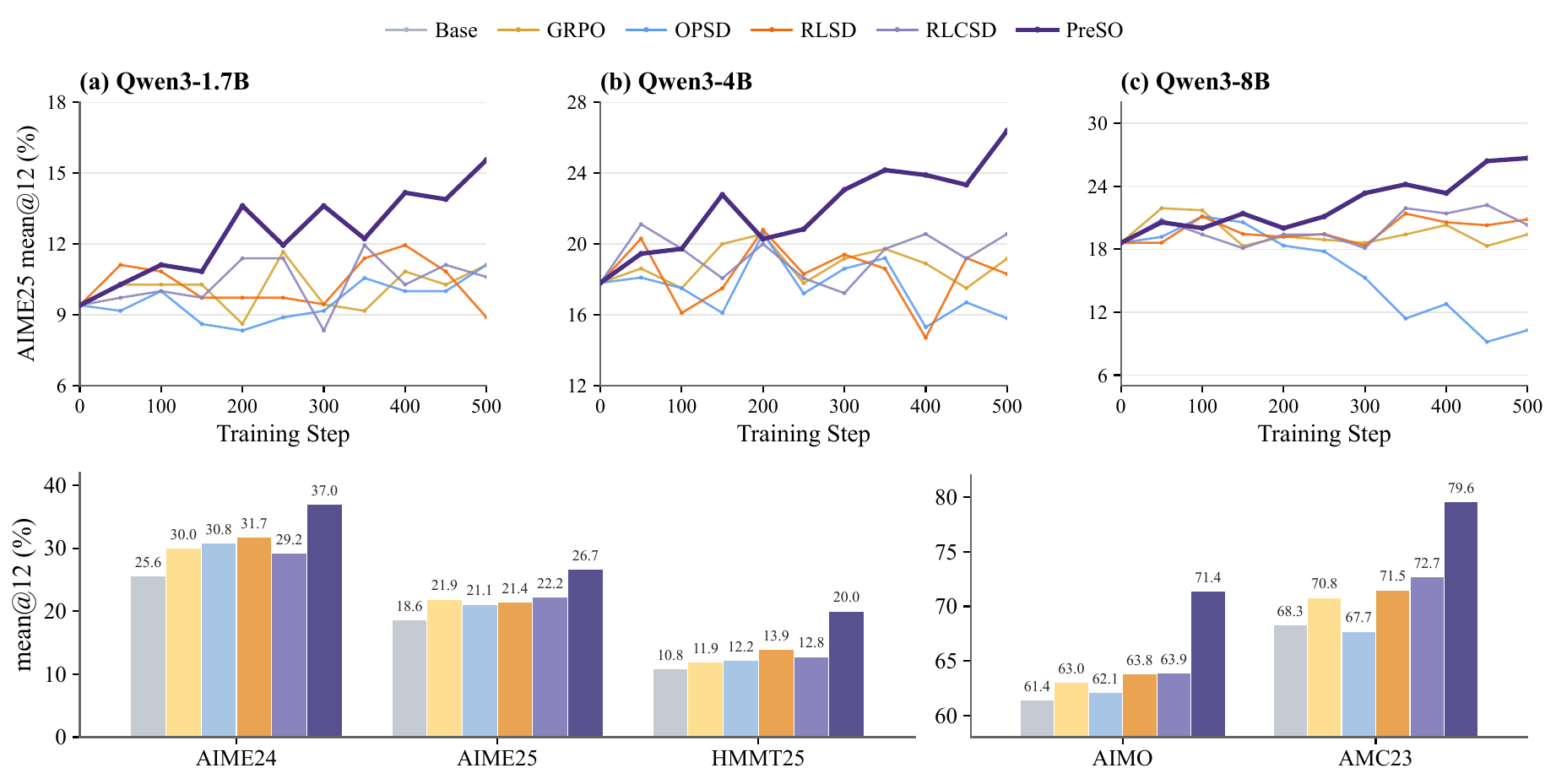}
\vspace{-20pt}
\caption{\textbf{Overall effectiveness of \ours{}.}
\textbf{Top:} Training dynamics across model scales. \ours{} maintains stable training curves throughout, with the advantage widening as the model scales.
\textbf{Bottom:} Performance comparison on Qwen3-8B. \ours{} consistently outperforms all baselines on every benchmark, reaching $79.6$ on AMC23 and surpassing the strongest baseline by $6.9$ points.}
\label{fig:main}
\end{figure*}

\section{Introduction}

Reinforcement learning with verifiable rewards (RLVR) has become a central paradigm for training reasoning models~\citep{grpo, deepseek-r1}. However, RLVR typically derives its learning signal from a sequence-level outcome reward, making credit assignment challenging for long-horizon reasoning trajectories~\citep{lightman2024let, wang2024math}. On-policy self-distillation (OPSD)~\citep{opsd, opcd} offers a denser alternative: conditioned on privileged context such as a reference solution, the same model acts as a self-teacher and provides token-level supervision on student-generated rollouts. This approach can improve both reasoning performance and sample efficiency over RLVR baselines~\citep{sdpo,many_face}.

To make better use of privileged teachers, prior work has improved OPSD through token reweighting \citep{zhang2026tailoring, PW-OPSD}, privileged context design \citep{opcd,wang2026skill,atesd,zhao2026training}, and integration with RLVR \citep{sdar,trace,rlsd}, with recent efforts extending online distillation to the representation level \citep{phf}. Other approaches contrast the teacher against a weaker reference to preserve the student's own reasoning ability \citep{rlcsd, ogls}. These methods refine how privileged guidance is provided or matched, but leave a more basic question underexplored: \textbf{which parts of a privileged teacher's advantage should the student actually internalize?} A privileged teacher is stronger precisely because it accesses information the student cannot obtain at inference time, yet the changes this information induces are not uniformly worth internalizing: some reflect reasoning regularities a student could in principle recover once the privilege is removed, while others are shortcuts tied to the known answer, a specific reference trajectory, or stylistic preferences of the teacher itself, which improve the teacher's trajectory under privileged conditions but do not tell the student how to reason when no such reference is available. Without separating the two, the student is pushed to imitate a pattern that can only be reproduced when privileged information is present.

The key question is not how much additional information the teacher receives, but whether the resulting changes exhibit stable patterns across related problems. A change that recurs among problems with similar content or reasoning structure may reveal a shared solving cue that the student can learn to use independently. By contrast, changes tied to a particular answer or reference trajectory offer little consistent signal across training examples and may be difficult to reuse without privileged information. We therefore use cross-problem recurrence to prioritize potentially reusable guidance, and evaluate its value through the student's performance without privileged information.

To examine privileged changes across problems, we turn to the model's \textit{internal representations}. For each rollout prefix, we define the \textit{privileged shift} as the difference between hidden states obtained with and without privileged information under the same prompt template. This lets us aggregate and compare shifts across problems. To probe how the detail of privileged context shapes these shifts, we compare full \textit{solutions}, which expose the complete derivation, with compact \textit{hints}, which provide only the key strategy. Despite revealing more, solutions lead to a steady decline in downstream performance, whereas hints yield sustained gains (Figure~\ref{fig:rep_pi_analysis}a). Hint-induced shifts concentrate along fewer dominant directions (Figure~\ref{fig:rep_pi_analysis}b), with lower effective rank and greater directional consistency than solution-induced shifts (Figure~\ref{fig:rep_pi_analysis}c). This contrast motivates using the leading low-rank subspace of pooled hint-induced shifts as an operational proxy for cross-problem structure and projecting privileged corrections onto it before constructing the distillation target.

Guided by this analysis, we propose \textbf{\oursfull{} (\ours{})}, which restricts privileged guidance at both the source and the representation-level target. At the \textit{source}, \ours{} replaces full reference solutions with compact hints to limit dependence on solution-specific details. At the \textit{target}, it routes each problem to a content-defined cluster, estimates the leading low-rank subspace of pooled hint-induced shifts, and projects the privilege-induced correction onto that subspace before constructing the distillation target. After warm-up, \ours{} projects only the privilege-induced correction
and adds it back to the student's unprivileged representation, selecting
directions of additional guidance without projecting away the representation the student already has. This design limits both the detail available to the privileged teacher and the directions of the correction supplied to the student. On five competition-level mathematics benchmarks, \ours{} consistently outperforms strong baselines across Qwen3-1.7B, Qwen3-4B, and Qwen3-8B, improving average accuracy over the base model by $\mathbf{6.8}$, $\mathbf{9.0}$, and $\mathbf{10.0}$ points, respectively. Its gains persist through later training rather than appearing only at an early peak.

\section{\ours{}: Privileged Representation Isolation for OPSD}
\label{sec:method}

\subsection{Preliminaries}

\paragraph{On-Policy Self-Distillation (OPSD).}
OPSD~\citep{opsd,sdpo,opcd} has recently emerged as a promising paradigm for on-policy knowledge distillation, offering dense token-level supervision while avoiding the training--inference distribution mismatch that plagues off-policy KD~\citep{opd,minillm}. Given a rollout \(y \sim \pi_{\theta_s}(\cdot \mid x)\), OPSD constructs a privileged self-teacher by conditioning the same model on additional information $\xi$, and matches \(\pi_{\theta_s}(\cdot \mid x,y_{<t})\) to \(\pi_{\theta_s}(\cdot \mid x,\xi,y_{<t})\) at each position \(t\):
\begin{equation}
\mathcal{L}_{\mathrm{OPSD}}
=
\mathbb{E}_{x,\, y\sim\pi_{\theta_s}(\cdot\mid x)}\!\left[\frac{1}{|y|}
\sum_{t=1}^{|y|}
D_{\mathrm{KL}}\!\left(
\operatorname{sg}\!\left(\pi_{\theta_s}(\cdot \mid x, \xi, y_{<t})\right)
\,\Big\|\,
\pi_{\theta_s}(\cdot \mid x, y_{<t})
\right)\right],
\label{eq:opsd}
\end{equation}
where $\operatorname{sg}(\cdot)$ denotes the stop-gradient operator. Compared with standard RLVR methods that optimize sparse outcome rewards, OPSD provides per-token supervision that densely tracks how the privilege-conditioned distribution differs from the student's.

However, privileged information can induce changes with different degrees of cross-problem consistency. A change that recurs across related problems may provide a reusable solving cue, while one tied to a particular answer or reference trajectory may offer little consistent guidance. We therefore use recurrence as an empirical cue for prioritizing privileged changes and evaluate their value through performance without privilege. Comparing these changes requires a common space in which their effects across problems can be aligned and aggregated. The model's internal representations provide such a space, which we study next.

\paragraph{Representation Self-Distillation.}
Representation distillation supervises intermediate features or attention
patterns rather than relying only on output distributions~\citep{fitnets,tinybert,minilm}.
In our on-policy setting, hidden-state matching bypasses the LM head and
uses a stop-gradient target computed on each student-generated rollout
$\hat y\sim\pi_{\theta_s}(\cdot\mid x)$. Following OPRD~\citep{oprd},
we match the student and privileged self-teacher representations at each
position $t$ under the shared prefix $\hat y_{<t}$ with an MSE loss
between $\ell_2$-normalized states:
\begin{equation}
\mathcal{L}
=
\frac{1}{|\mathcal T|}
\sum_{(i,t)\in\mathcal T}
\frac{1}{d}
\big\| h_{\theta_s}(x_i, \hat y_{<t}) - \operatorname{sg}\!\left(h_{\theta_s}(x_i, \xi_i, \hat y_{<t})\right) \big\|_{2}^{2},
\label{eq:oprsd}
\end{equation}
where $h_{\theta_s}(x_i, \hat y_{<t})$ and $h_{\theta_s}(x_i, \xi_i,
\hat y_{<t})$ are the final-layer student and self-teacher hidden
states at position $t$, $\mathcal{T}$ collects the valid supervised
positions, and $d$ is the hidden dimension. The supervision thus
depends entirely on how $\xi_i$ shifts the student's hidden states
along its own rollouts.

We make this concrete through the \emph{privileged shift}. At each
rollout position $t$, conditioning on $\xi$ moves the model's hidden
state, inducing a representation shift
$\Delta_t = h_{\theta_s}(x,\xi,\hat{y}_{<t}) - h_{\theta_s}(x,\hat{y}_{<t}) \in \mathbb{R}^d$,
which the LM head then maps onto the next-token distribution. Since
every $\Delta_t$ lives in the same $\mathbb{R}^d$, shifts from
different problems share one coordinate system in which their
directions can be compared and their recurring structure measured.
We examine this geometry next.


\begin{figure}[t]
\centering
\includegraphics[width=\linewidth]{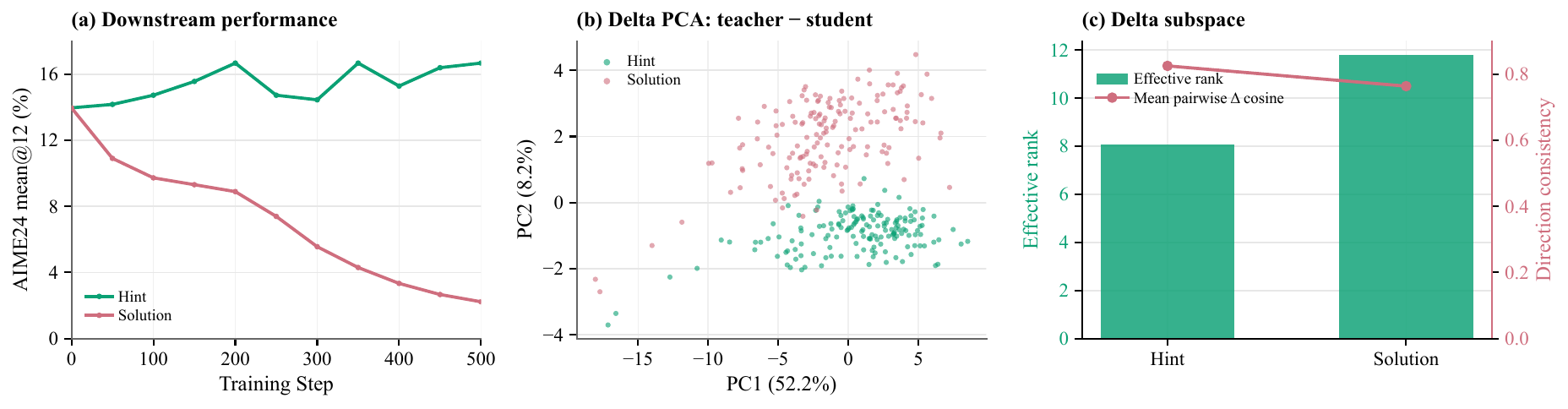}
\vspace{-12pt}
\caption{\textbf{Privileged-information geometry.} (a) Distilling from hints sustains downstream accuracy over training, whereas full solutions degrade steadily. (b) PCA of the privileged shifts separates solutions into a broad cloud that leaks into $\mathrm{PC2}$, while hints stay concentrated along $\mathrm{PC1}$ with a common orientation. (c) The hint-induced shift attains a lower effective rank and higher directional consistency, whereas the full solution spreads over a higher-rank, less consistent structure.}
\label{fig:rep_pi_analysis}
\vspace{-5pt}
\end{figure}

\subsection{How Privileged-Information Granularity Shapes the Shift}
\label{sec:pi-analysis}

A direct way to shape the privileged shift is to control what the self-teacher sees. Prior work~\citep{opcd,wang2026skill,atesd,zhao2026training} has shown that the form of privileged information can substantially affect distillation. Here we study how its granularity changes both the cross-problem geometry of $\Delta_t$ and the downstream effect of distilling from it. We compare a ground-truth \emph{solution}, which exposes the complete derivation and final answer, with a compact \emph{hint}, which provides a key sub-step or partial derivation without disclosing the answer.

More detailed privilege does not necessarily yield better supervision. In this comparison, hints raise downstream accuracy from roughly $14\%$ to above $16\%$ and sustain that gain, whereas solutions fall to $2\text{--}3\%$ (Figure~\ref{fig:rep_pi_analysis}a). To examine the corresponding geometry, we stack per-position shifts across problems and project them onto their leading principal components. Hint-induced shifts occupy a narrow, consistently oriented region along $\mathrm{PC1}$, which explains $52.2\%$ of the pooled variance, while solution-induced shifts spread more broadly toward $\mathrm{PC2}$, which explains $8.2\%$ (Figure~\ref{fig:rep_pi_analysis}b). Hint-induced shifts also have lower effective rank and greater directional consistency than solution-induced shifts (Figure~\ref{fig:rep_pi_analysis}c). In this comparison, the more concentrated and consistent shifts accompany the more sustained downstream gain. We therefore use the leading low-rank subspace of pooled hint-induced shifts as an operational proxy for recurring structure in Section~\ref{sec:low-rank}.

\subsection{Privileged Subspace Projection}
\label{sec:low-rank}
Guided by the geometry in Section~\ref{sec:pi-analysis}, we use the leading low-rank subspace of pooled privileged shifts as an operational proxy for cross-problem structure. \ours{} projects the privilege-induced shift onto this subspace before adding it to the unprivileged student representation to form a distillation target. We next describe how the subspace is estimated and applied.
\paragraph{Estimating the privileged subspace.}
During the first \(W\) optimization steps, we collect rollouts from the evolving on-policy student \(\pi_{\theta_s}\). For each visited prefix \((x_i,\hat y_{<t})\), we evaluate the parameter-shared self-teacher twice under an identical template that differs only in whether the privileged information \(\xi_i\) is exposed, giving the privileged shift:
\begin{equation}
\Delta_{i,t}=h_{\theta_s}(x_i,\xi_i,\hat y_{<t})-h_{\theta_s}(x_i,\varnothing,\hat y_{<t}),
\label{eq:pi-shift-method}
\end{equation}
where \(\varnothing\) denotes the same template with an empty privileged field. Holding both the template and the rollout prefix isolates the representational effect of privileged information. We stack the collected shifts row-wise into \(D\in\mathbb{R}^{M\times d}\) and compute an \textit{uncentered} rank-\(r\) SVD:
\begin{equation}
D \approx U\Sigma V^\top,\qquad V\in\mathbb{R}^{d\times r},
\label{eq:svd}
\end{equation}
whose right singular vectors $V$ span the $r$ dominant energy directions of the
pooled shifts in representation space. The retained directions define the orthogonal projector $P=VV^\top$. By construction, this projector preserves the largest share of observed shift energy among rank-$r$ projectors. Motivated by the geometry in Section~\ref{sec:pi-analysis}, we use their span as an empirical proxy for cross-problem structure and apply $P$ to restrict the directions of subsequent privileged corrections.

\begin{figure}[t]
\centering
\includegraphics[width=\linewidth]{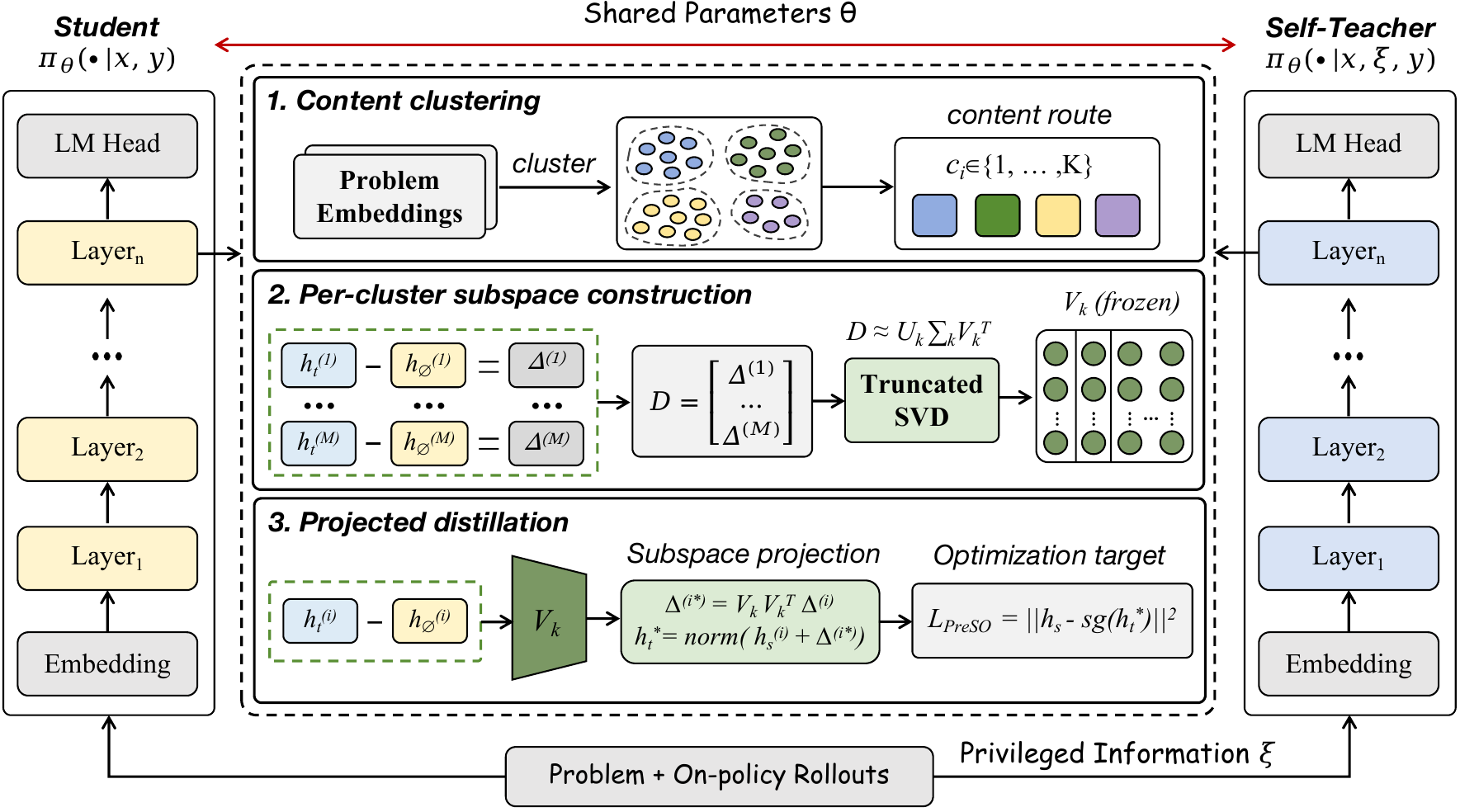}
\caption{\textbf{Overview of \ours{}.} A compact hint conditions the self-teacher. During warm-up, \ours{} routes problems by content and estimates a leading low-rank subspace from privileged shifts pooled within each cluster. It then freezes these subspaces and, during subsequent on-policy distillation, projects each privileged shift onto its assigned subspace before constructing the representation-level target. The selected subspace serves as an operational proxy for cross-problem structure.}
\label{fig:oprsd}
\vspace{-5pt}
\end{figure}

\paragraph{Content-based subspace routing.}
The construction above yields a single global subspace. Across heterogeneous problem families, however, the recurring directions themselves may vary, so one global subspace risks conflating incompatible reasoning patterns. We therefore route each problem to a content-specific subspace using only privilege-free, student-visible information. For problem \(x_i\), we read its last prompt-position representation \(\ell_i\) from the frozen \(\theta_0\) and normalize it:
\begin{equation}
e_i
=
\operatorname{norm}\!\left(
h_{\theta_0}^{(L)}(x_i)_{\ell_i}
\right),
\label{eq:content-embedding}
\end{equation}
where \(L\) is the final transformer layer. Reading from the frozen \(\theta_0\) anchors routing to a  representation space, keeping each problem's assignment independent of the evolving policy and consistent across training. We then run \(K\)-means on \(\{e_i\}_{i=1}^{N}\) and assign each to its nearest centroid:
\begin{equation}
c_i
=
\arg\min_{c\in\{1,\ldots,K\}}
\|e_i-\mu_c\|_2^2,
\label{eq:content-cluster}
\end{equation}
where \(\mu_c\) is the cluster centroid. The route is computed once before training and frozen, so all rollouts of a problem share one assignment, and since it depends on neither privileged information nor sampled responses, it introduces no leakage or rollout dependence. During warm-up, each shift updates only the second moment of its assigned cluster, giving one projector \(P_c=V_cV_c^{\top}\) per route.

\subsection{The Optimization Objective}
\label{sec:objective}
Training proceeds in two stages. For the first \(W\) steps we match the full representation while accumulating the privileged shifts \(\Delta_{i,t}\) of Eq.~\ref{eq:pi-shift-method}, each measured from the parameter-shared self-teacher so that it reflects the net effect of the privileged information rather than a template artifact. We then construct and freeze the cluster projectors \(\{P_c\}_{c=1}^{K}\) and switch to projected distillation, adding the projected shift back onto the student anchor \(h_{\theta_s}(x_i,\hat y_{<t})\). The stage-dependent target is:
\begin{equation}
h_{i,t}^{\star(s)} =
\begin{cases}
\operatorname{sg}\!\left[
h_{\theta_s}(x_i,\xi_i,\hat y_{<t})
\right],
& s\leq W,\\[1mm]
\operatorname{sg}\!\left[
\operatorname{norm}\!\left(
h_{\theta_s}(x_i,\hat y_{<t})
+ P_{c_i}\Delta_{i,t}
\right)
\right],
& s>W.
\end{cases}
\label{eq:stage-target}
\end{equation}
During warm-up, the student matches the full privileged representation.
Afterwards, before normalization, the target combines the teacher's
representation in the selected subspace with the student's unprivileged
representation in its orthogonal complement. We then normalize the
target and compute squared-error loss over the full hidden vector:
\begin{equation}
\mathcal L_{\ours{}}
=
\frac{\lambda}{|\mathcal T|}
\sum_{(i,t)\in\mathcal T}
\frac{1}{d}
\left\|
h_{\theta_s}(x_i,\hat y_{<t})
-
h_{i,t}^{\star(s)}
\right\|_2^2,
\label{eq:PreSO-loss}
\end{equation}
where $\mathcal T$ contains the valid supervised positions, $d$ is the hidden dimension, and $\lambda$ is a scalar coefficient on the loss. Student and self-teacher states are $\ell_2$-normalized before computing each shift, and the projected target is normalized before evaluating the loss. This gives the teacher-induced correction a low-rank metric interpretation~\citep{weinberger2009distance}, while the loss still compares full normalized representations. Algorithm \ref{alg:PreSO} shows the pseudo-code of \ours{}.

\begin{algorithm}[t]
\caption{\ours{}: \oursfull{}}
\label{alg:PreSO}
\begin{algorithmic}[1]
\Require Initial policy \(\pi_{\theta}\), training set
\(\mathcal D=\{(x_i,\xi_i)\}_{i=1}^{N}\), clusters \(K\),
rank \(r\), warm-up steps \(W\)

\Statex \hspace{0.5em}{\color{blue}\(\triangleright\)
\textbf{Problem Content Routing}}
\State Extract \(\{e_i\}\), assign each \(x_i\) to \(c_i\) via \(K\)-means,
and initialize the route statistics \(\{C_c\}_{c=1}^{K}\)

\Statex \hspace{0.5em}{\color{blue}\(\triangleright\)
\textbf{On-Policy Subspace Estimation}}
\For{\(s=1,\ldots,W\)}
    \State Sample on-policy rollouts from \(\pi_\theta\), compute
    \(\Delta_{i,t}\) by Eq.~(\ref{eq:pi-shift-method}), and update \(C_{c_i}\)
    \State Optimize the warm-up branch of
    Eqs.~(\ref{eq:stage-target})--(\ref{eq:PreSO-loss})
\EndFor
\State Extract the top-\(r\) eigenspaces \(\{V_c\}_{c=1}^{K}\) and freeze
\(P_c\leftarrow V_cV_c^\top\) for each route

\Statex \hspace{0.5em}{\color{blue}\(\triangleright\)
\textbf{Projected Self-Distillation}}
\Repeat
    \State Sample on-policy rollouts from \(\pi_\theta\), compute
    \(\Delta_{i,t}\), and retrieve \(P_{c_i}\)
    \State Construct projected targets by Eq.~(\ref{eq:stage-target}), compute
\(\mathcal L_{\ours{}}\) by Eq.~(\ref{eq:PreSO-loss}), and update \(\theta\)
\Until{converged}
\end{algorithmic}
\end{algorithm}

\section{Experiments}

\subsection{Experimental Setup}

\paragraph{Models and Datasets.}
We conduct experiments with Qwen3-1.7B, Qwen3-4B and Qwen3-8B~\citep{qwen3}, disabling thinking mode throughout training and evaluation; results under the thinking mode are reported in Appendix~\ref{app:nothink-think}. To test robustness across training data, we use two training sets, DeepMath-103K~\citep{deepmath} and Open-R1~\citep{luffy}. The main tables report results trained on DeepMath-103K, and the Open-R1 results are given in Appendix~\ref{app:openr1_result}. For DeepMath, following RLCSD~\citep{rlcsd}, we assign difficulty levels 5--7 to the 1.7B model, levels 6--8 to the 4B model, and levels 7--10 to the 8B model. Each model is trained for 500 optimization steps. We evaluate on five competition-level mathematical reasoning benchmarks: AIME24~\citep{aime24}, AIME25~\citep{aime25}, AIMO~\citep{aimo}, HMMT25~\citep{hmmt25}, and AMC23~\citep{amc23}. For efficiency, the privileged hints are annotated offline with GPT-5.5 before training, and Appendix~\ref{app:self-hint} reports a self-hint variant.

\paragraph{Baselines.}
We compare \ours{} against four representative baselines.
(1) \textbf{GRPO}~\citep{grpo}, an RLVR baseline that optimizes the policy with
group-normalized advantages from verifiable outcome rewards.
(2) \textbf{OPSD}~\citep{opsd}, an on-policy self-distillation method in which one model
acts as both privileged teacher and student and minimizes their per-token divergence over
its own rollouts.
(3) \textbf{RLSD}~\citep{rlsd}, a hybrid method that sizes each policy update from the
teacher-student token-level discrepancy while taking its direction from RLVR outcome feedback.
(4) \textbf{RLCSD}~\citep{rlcsd}, a contrastive self-distillation method that conditions the teacher on successful and unsuccessful on-policy sibling solutions and uses their likelihood contrast to refine the policy advantages on task-relevant tokens. For a fair comparison, OPSD and RLSD in the main results receive the same hints as privileged information, and results using the full solution are in Appendix~\ref{app:solution-pi}. Implementation details are in Appendix~\ref{app:training-details}.

\begin{table*}[!t]
\vspace{0pt}
\caption{Experimental results (\%) on mathematical reasoning benchmarks. For each method and benchmark, we report the highest mean@12 across evaluated checkpoints and their average (Avg.). Last-checkpoint results are reported in Appendix~\ref{app:last-ckpt}. Within each base-model group, the best and second-best scores are highlighted in \textbf{bold} and \underline{underlined}, respectively.}
\vspace{0pt}
\centering
\small
\renewcommand{\arraystretch}{1.20}
\setlength{\tabcolsep}{5pt}

\begin{tabularx}{0.96\textwidth}{l|*{6}{Y}}
\toprule
\textbf{Method}
& \textbf{AIME24}
& \textbf{AIME25}
& \textbf{AIMO}
& \textbf{HMMT25}
& \textbf{AMC23}
& \textbf{Avg.} \\
\midrule

\textbf{Qwen3-8B}
& 25.6 \basex{0.0}
& 18.6 \basex{0.0}
& 61.4 \basex{0.0}
& 10.8 \basex{0.0}
& 68.3 \basex{0.0}
& 36.9 \basex{0.0} \\

\hspace*{1em}+ GRPO
& 30.0 \uprise{4.4}
& 21.9 \uprise{3.3}
& 63.0 \uprise{1.6}
& 11.9 \uprise{1.1}
& 70.8 \uprise{2.5}
& 39.5 \uprise{2.6} \\

\hspace*{1em}+ OPSD
& 30.8 \uprise{5.2}
& 21.1 \uprise{2.5}
& 62.1 \uprise{0.7}
& 12.2 \uprise{1.4}
& 67.7 \down{0.6}
& 38.8 \uprise{1.9} \\

\hspace*{1em}+ RLSD
& \underline{31.7} \uprise{6.1}
& 21.4 \uprise{2.8}
& 63.8 \uprise{2.4}
& \underline{13.9} \uprise{3.1}
& 71.5 \uprise{3.2}
& \underline{40.5} \uprise{3.6} \\

\hspace*{1em}+ RLCSD
& 29.2 \uprise{3.6}
& \underline{22.2} \uprise{3.6}
& \underline{63.9} \uprise{2.5}
& 12.8 \uprise{2.0}
& \underline{72.7} \uprise{4.4}
& 40.2 \uprise{3.3} \\

\rowcolor{blue!10}
\hspace*{1em}+ \textbf{\ours{}}
& \textbf{37.0} \uprise{11.4}
& \textbf{26.7} \uprise{8.1}
& \textbf{71.4} \uprise{10.0}
& \textbf{20.0} \uprise{9.2}
& \textbf{79.6} \uprise{11.3}
& \textbf{46.9} \uprise{10.0} \\

\midrule

\textbf{Qwen3-4B}
& 23.3 \basex{0.0}
& 17.8 \basex{0.0}
& 61.2 \basex{0.0}
& 11.4 \basex{0.0}
& 66.5 \basex{0.0}
& 36.0 \basex{0.0} \\

\hspace*{1em}+ GRPO
& 25.6 \uprise{2.3}
& 20.6 \uprise{2.8}
& 63.0 \uprise{1.8}
& 13.1 \uprise{1.7}
& 68.5 \uprise{2.0}
& 38.2 \uprise{2.2} \\

\hspace*{1em}+ OPSD
& 25.3 \uprise{2.0}
& 20.6 \uprise{2.8}
& 60.2 \down{1.0}
& 12.5 \uprise{1.1}
& \underline{71.3} \uprise{4.8}
& 38.0 \uprise{2.0} \\

\hspace*{1em}+ RLSD
& \underline{27.5} \uprise{4.2}
& 20.8 \uprise{3.0}
& \underline{63.1} \uprise{1.9}
& \underline{13.3} \uprise{1.9}
& 70.4 \uprise{3.9}
& \underline{39.0} \uprise{3.0} \\

\hspace*{1em}+ RLCSD
& \underline{27.5} \uprise{4.2}
& \underline{21.1} \uprise{3.3}
& 61.9 \uprise{0.7}
& 13.1 \uprise{1.7}
& 70.8 \uprise{4.3}
& 38.9 \uprise{2.9} \\

\rowcolor{blue!10}
\hspace*{1em}+ \textbf{PreSO}
& \textbf{37.2} \uprise{13.9}
& \textbf{26.4} \uprise{8.6}
& \textbf{67.9} \uprise{6.7}
& \textbf{18.1} \uprise{6.7}
& \textbf{75.6} \uprise{9.1}
& \textbf{45.0} \uprise{9.0} \\

\midrule

\textbf{Qwen3-1.7B}
& 12.8 \basex{0.0}
& 9.4 \basex{0.0}
& 40.3 \basex{0.0}
& 5.6 \basex{0.0}
& 44.8 \basex{0.0}
& 22.6 \basex{0.0} \\

\hspace*{1em}+ GRPO
& 13.6 \uprise{0.8}
& 11.7 \uprise{2.3}
& 41.8 \uprise{1.5}
& 6.4 \uprise{0.8}
& 47.5 \uprise{2.7}
& 24.2 \uprise{1.6} \\

\hspace*{1em}+ OPSD
& 16.1 \uprise{3.3}
& 11.1 \uprise{1.7}
& 43.0 \uprise{2.7}
& 6.7 \uprise{1.1}
& 47.7 \uprise{2.9}
& 24.9 \uprise{2.3} \\

\hspace*{1em}+ RLSD
& \underline{16.4} \uprise{3.6}
& \underline{11.9} \uprise{2.5}
& \underline{44.0} \uprise{3.7}
& \underline{6.9} \uprise{1.3}
& 47.9 \uprise{3.1}
& \underline{25.4} \uprise{2.8} \\

\hspace*{1em}+ RLCSD
& 15.6 \uprise{2.8}
& \underline{11.9} \uprise{2.5}
& 42.5 \uprise{2.2}
& 6.7 \uprise{1.1}
& \underline{48.1} \uprise{3.3}
& 25.0 \uprise{2.4} \\

\rowcolor{blue!10}
\hspace*{1em}+ \textbf{PreSO}
& \textbf{18.9} \uprise{6.1}
& \textbf{15.6} \uprise{6.2}
& \textbf{48.2} \uprise{7.9}
& \textbf{9.7} \uprise{4.1}
& \textbf{54.6} \uprise{9.8}
& \textbf{29.4} \uprise{6.8} \\

\bottomrule
\end{tabularx}
\label{tb:main}
\vspace{-5pt}
\end{table*}

\subsection{Main results}
\paragraph{\ours{} outperforms existing baselines.}
Table~\ref{tb:main} shows that \ours{} achieves the strongest results across all model scales and benchmarks. On Qwen3-8B, it raises average accuracy from $36.9$ to $\mathbf{46.9}$, gaining $\mathbf{+10.0}$ points over the base model and $\mathbf{+6.4}$ over the strongest baseline, RLSD. Average gains over the base model reach $\mathbf{+9.0}$ points on Qwen3-4B and $\mathbf{+6.8}$ on Qwen3-1.7B. On AIME24 and HMMT25, Qwen3-8B with \ours{} scores $\mathbf{37.0}$ and $\mathbf{20.0}$, compared with $31.7$ and $13.9$ for the strongest baseline. These gains extend across benchmarks and increase with model scale. Whereas OPSD matches privileged behavior at the token level and RLCSD uses a token-level contrastive correction, \ours{} constructs representation-level targets from projected shifts.

\paragraph{\ours{} remains stable throughout training.}
Figure~\ref{fig:main} shows that OPSD peaks early and then declines, particularly at larger scales, whereas \ours{} sustains its gains through the end of training. Table~\ref{tb:last_ckpt} confirms that \ours{} attains the highest final-checkpoint average on all three backbones, with final scores close to their peaks. Its distillation loss directly matches normalized hidden states without passing through the LM head's vocabulary softmax. After warm-up, \ours{} projects each privileged shift onto an estimated per-cluster subspace before constructing the representation-level target. Subspace ablations show that matching the full shift or projecting onto random or tail directions yields weaker performance. These comparisons suggest that restricting the privileged correction to selected leading directions contributes to the sustained gain.

\subsection{Analysis}
\label{sec:analysis}

\begin{figure}[t]
\vspace{3pt}
\centering
\includegraphics[width=\linewidth]{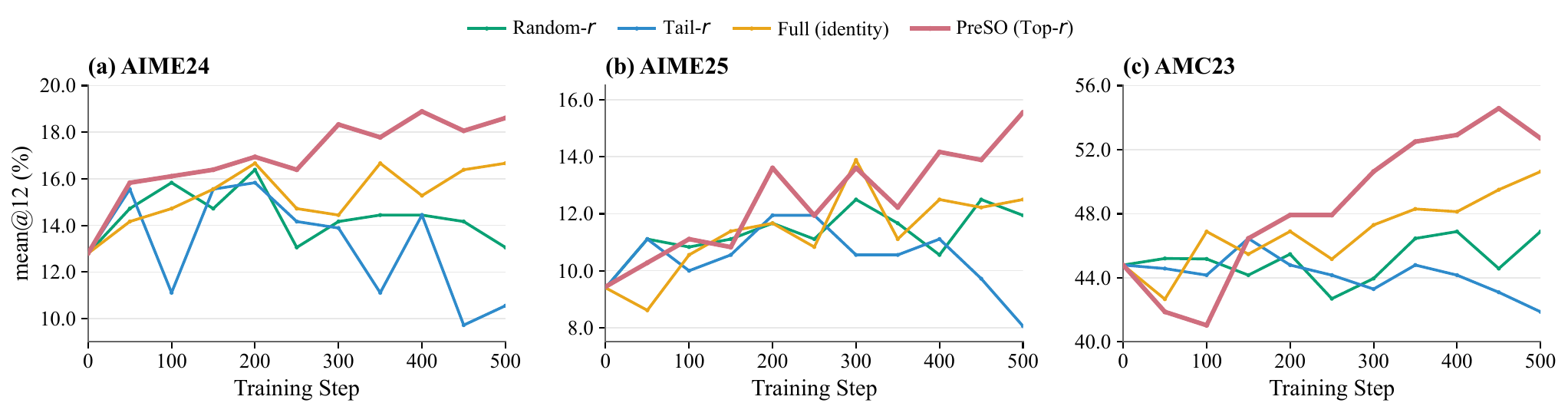}
\vspace{-13pt}
\caption{\textbf{Effect of the projection subspace.} Mean@12 over training on AIME24, AIME25, and AMC23 with Qwen3-1.7B. Top-$r$, Random-$r$, and Tail-$r$ project the privileged shift onto leading, random, and lowest-energy rank-$r$ subspaces, respectively. Full uses the unprojected shift.}
\label{fig:subspace-ablation}
\vspace{-5pt}
\end{figure}

\paragraph{The effect of the low-rank subspace.}
We examine whether \ours{} benefits from a low-rank constraint alone or from the particular directions it retains. At rank $r$, we compare its top-$r$ subspace with random-$r$ and bottom-$r$ subspaces, and also evaluate the full shift without projection. Figure~\ref{fig:subspace-ablation} shows that all three alternatives finish below \ours{}. The full-shift variant remains competitive for much of training but plateaus lower. Random-$r$ improves slowly, while Tail-$r$ is the least stable and drifts toward the base level late in training. \ours{} retains the leading directions of the pooled shifts, our operational proxy for cross-problem structure, and achieves the highest final accuracy. Restricting the correction to $r$ directions may make it easier to learn, but the same-rank comparisons show that reducing dimension alone does not explain the gain. Top-$r$ also remains ahead when Random-$r$ and Tail-$r$ are rescaled to match its retained shift norm (Appendix~\ref{app:energy-matched}), supporting a role for the selected directions beyond differences in correction magnitude. Controlled hint perturbations show greater subspace overlap and more
sustained gains for correct hints than for length- and norm-matched
wrong or cross-problem hints (Appendix~\ref{app:subspace-probe}). A concept-only source,
however, also produces concentrated, directionally consistent shifts
but underperforms hints, indicating that this geometry alone does
not establish useful guidance (Appendix~\ref{app:concept}).

\begin{figure}[t]
\vspace{-12pt}
\centering
\includegraphics[width=\linewidth]{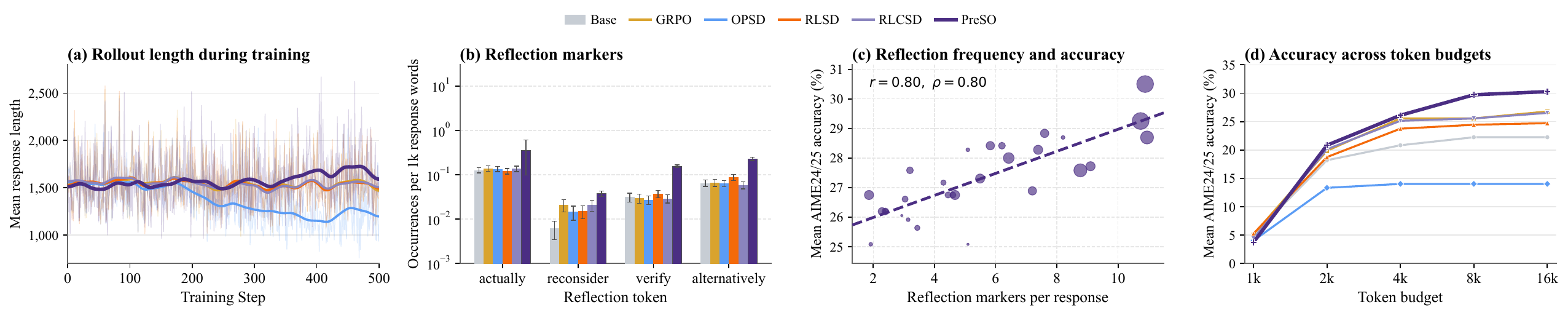}
\vspace{-12pt}
\caption{\textbf{\ours{} elicits productive, exploratory reasoning.} (a)~Mean rollout length over training.
(b)~Reflection-marker density per 1k words. (c)~Across checkpoints, marker density
versus accuracy ($r=0.80$); bubble size is the successful self-correction rate.
(d)~Accuracy across token budgets at the final checkpoint (step 500). Statistics use Qwen3-8B on AIME24/25.}
\label{fig:reflection}
\vspace{-15pt}
\end{figure}

\paragraph{\ours{} sustains reflective rollouts.}
Beyond accuracy, we examine generation behavior during training on Qwen3-8B (Fig.~\ref{fig:reflection}). \ours{} maintains long rollouts (Fig.~\ref{fig:reflection}a), with a higher density of reflection markers such as \emph{actually}, \emph{reconsider}, and \emph{alternatively} than any baseline (Fig.~\ref{fig:reflection}b). Across checkpoints, marker density correlates with AIME24/25 accuracy ($r=0.80$), and higher-density checkpoints more often show explicit error detection and revision (Fig.~\ref{fig:reflection}c). At fixed generation budgets, \ours{} leads from 2k to 16k tokens, reaching by 4k the accuracy that the strongest baseline attains only with a larger budget (Fig.~\ref{fig:reflection}d). Under equal wall-clock training budgets, \ours{} also leads every baseline on all five benchmarks at both evaluated scales (Appendix~\ref{app:efficiency}).

\subsection{Ablation Studies}
\label{sec:ablation}

\paragraph{Ablation on the source-side abstraction of privileged information.}
We first replace compact hints with full solutions while keeping the rest of \ours{} unchanged (w/o PI abstraction). On Qwen3-8B, this variant trails \ours{} from the outset and declines most sharply, ending $15.3$, $10.6$, and $12.9$ points lower on AIME24, AIME25, and AMC23 at the last checkpoint. These are the largest gaps among the ablations. To test whether the loss stems solely from the subspace estimated from solution-induced shifts, we also estimate the subspace from hints while retaining solutions as privileged context (w/o PI abs. (hint subspace)). This recovers only part of the gap and still trails \ours{} by $12.6$ points on AIME24. The same pattern holds on Qwen3-1.7B. Together, these results show that compact hints contribute beyond their effect on subspace estimation. They are consistent with source-side abstraction reducing potential interference from solution-specific details, which target-side projection alone does not fully compensate for.

\begin{table}[t]
\caption{\textbf{Ablation study on \ours{}.}
Accuracy (\%) on AIME24/AIME25/AMC23 at the best and last checkpoints.
Subscripts show each variant's gap to \ours{}.}
\centering
\small
\renewcommand{\arraystretch}{1.0}

\resizebox{\linewidth}{!}{%
\begin{tabular}{lcccccc}
\toprule
\multirow{2}{*}{\textbf{Method}}
& \multicolumn{3}{c}{\textbf{Best ckpt}}
& \multicolumn{3}{c}{\textbf{Last ckpt}} \\
\cmidrule(lr){2-4}
\cmidrule(lr){5-7}
& \textbf{AIME24}
& \textbf{AIME25}
& \textbf{AMC23}
& \textbf{AIME24}
& \textbf{AIME25}
& \textbf{AMC23} \\
\midrule

\textbf{\ours{} (Qwen3-8B)}
& \textbf{37.0}
& \textbf{26.7}
& \textbf{79.6}
& \textbf{37.0}
& \textbf{26.7}
& \textbf{79.4} \\

\hspace*{0.5em}w/o PI abstraction
& 28.6 \down{8.4}
& 19.7 \down{7.0}
& 69.4 \down{10.2}
& 21.7 \down{15.3}
& 16.1 \down{10.6}
& 66.5 \down{12.9} \\

\hspace*{0.5em}w/o PI abs. (hint subspace)
& 32.8 \down{4.2}
& 20.3 \down{6.4}
& 71.3 \down{8.3}
& 24.4 \down{12.6}
& 16.4 \down{10.3}
& 68.3 \down{11.1} \\

\hspace*{0.5em}w/o low-rank projection
& 33.6 \down{3.4}
& 22.8 \down{3.9}
& 73.8 \down{5.8}
& 33.6 \down{3.4}
& 22.5 \down{4.2}
& 73.8 \down{5.6} \\

\hspace*{0.5em}w/o cluster routing
& 34.4 \down{2.6}
& 23.3 \down{3.4}
& 75.0 \down{4.6}
& 34.4 \down{2.6}
& 21.1 \down{5.6}
& 73.1 \down{6.3} \\

\midrule

\textbf{\ours{} (Qwen3-1.7B)}
& 18.9
& \textbf{15.6}
& \textbf{54.6}
& 18.6
& \textbf{15.6}
& \textbf{52.7} \\

\hspace*{0.5em}w/o PI abstraction
& 13.3 \down{5.6}
& 8.9 \down{6.7}
& 44.0 \down{10.6}
& 6.1 \down{12.5}
& 5.6 \down{10.0}
& 40.6 \down{12.1} \\

\hspace*{0.5em}w/o PI abs. (hint subspace)
& 15.0 \down{3.9}
& 11.7 \down{3.9}
& 46.7 \down{7.9}
& 11.1 \down{7.5}
& 8.9 \down{6.7}
& 44.6 \down{8.1} \\

\hspace*{0.5em}w/o low-rank projection
& 16.7 \down{2.2}
& 13.6 \down{2.0}
& 50.6 \down{4.0}
& 16.7 \down{1.9}
& 12.5 \down{3.1}
& 50.6 \down{2.1} \\

\hspace*{0.5em}w/o cluster routing
& \textbf{19.1} \uprise{0.2}
& 14.2 \down{1.4}
& 52.9 \down{1.7}
& \textbf{19.1} \uprise{0.5}
& 12.5 \down{3.1}
& \textbf{52.7} \basex{0.0} \\

\bottomrule
\end{tabular}%
}
\label{tab:ablation}
\end{table}

\begin{figure}[t]
\vspace{-12pt}
\centering
\includegraphics[width=\linewidth]{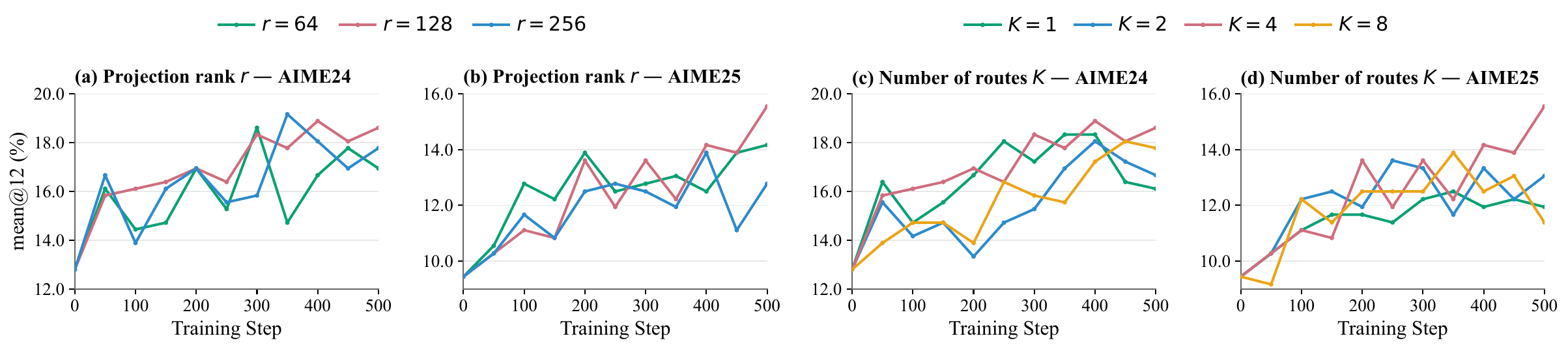}
\vspace{-12pt}
\caption{\textbf{Ablation on the projection rank $r$ and the number of routes $K$.}
mean@12 over training steps on AIME24 and AIME25 with Qwen3-1.7B. Panels (a) and (b) vary the low-rank projection rank $r \in \{64, 128, 256\}$, and panels (c) and (d) vary the number of routes $K \in \{1, 2, 4, 8\}$.}
\label{fig:ablation}
\vspace{-5pt}
\end{figure}

\paragraph{Ablation on the low-rank projection and cluster routing.}
Holding the hint source fixed, we ablate the two target-side components. On Qwen3-8B, using the full privileged shift instead of a projected correction (w/o low-rank) lowers best-checkpoint accuracy by $3.4$ and $3.9$ points on AIME24 and AIME25. Removing cluster routing (w/o cluster) has a smaller effect at the peak, but its gap to \ours{} on AIME25 widens from $3.4$ points at the best checkpoint to $5.6$ at the last. Similar trends hold on Qwen3-1.7B. These results suggest complementary contributions from projection and routing, with routing particularly relevant to late-stage performance. Figure~\ref{fig:ablation} varies the projection rank $r\in\{64,128,256\}$ and route count $K\in\{1,2,4,8\}$ around the defaults of $128$ and $4$. Rank $64$ retains fewer directions and trails the default, while rank $256$ includes additional lower-energy directions and loses its early advantage later in training. Using one route forces all problems to share a subspace, whereas eight routes leave fewer warm-up shifts in each cluster and yield less stable training. Among the tested settings, $r=128$ and $K=4$ perform best. More supplementary experiments and further analyses are provided in the Appendix~\ref{app:more_experiments} and Appendix~\ref{app:more_analysis}.

\section{Related Work}

\paragraph{On-policy self-distillation.}
On-policy self-distillation (OPSD) conditions the current policy on privileged information or feedback to obtain a self-teacher that provides dense token-level supervision on student-generated rollouts, avoiding the distribution mismatch of off-policy teachers~\citep{opd,opsd,sdpo}. Prior work has improved this supervision through privileged context design~\citep{opcd,sdpo-u}, token reweighting~\citep{PW-OPSD,zhang2026tailoring}, and integration with RLVR~\citep{rlsd,sdar,srpo}. Other approaches selectively correct the teacher signal to reduce style- or reference-dependent effects~\citep{rlcsd,ogls,purified_opsd}. These methods primarily operate on token-level distributions or logits. We instead study how privileged information shifts hidden representations across problems. \ours{} uses compact hints to shape these shifts and estimates a leading low-rank subspace from shifts pooled across related problems as an empirical proxy for cross-problem structure. It then projects the privilege-induced correction onto this subspace before constructing the representation-level target.

\paragraph{Hidden representation distillation.}
Representation distillation complements output-distribution matching by aligning hidden states~\citep{fitnets,tinybert,mobilebert} or relations among representations~\citep{rkd,crd}. Recent work has brought such supervision to on-policy rollouts. OPRD~\citep{oprd} aligns student and teacher hidden states on shared rollouts, while PHF~\citep{phf} studies privileged self-distillation through hidden-state transitions and trajectory geometry. Our use of representations serves an additional purpose: they provide a common space for comparing the changes induced by privileged information across problems and assessing whether dominant directions recur among related problems. We analyze these shifts to examine how their directional structure relates to downstream learning. Guided by this analysis, \ours{} uses the leading subspace of pooled shifts as an empirical proxy for recurring structure and projects the privilege-induced correction before constructing its distillation target.

\section{Conclusion}
In this paper, we revisit on-policy self-distillation by asking which privileged changes can help a student reason without privileged information. We use cross-problem recurrence as an empirical cue and examine privilege-induced representation shifts across problems. Hint-induced shifts exhibit lower effective rank and greater directional consistency than solution-induced shifts, alongside more sustained downstream gains. Motivated by this association, we propose \ours{}, which uses compact hints at the source and projects subsequent privilege-induced corrections onto per-cluster leading subspaces before constructing representation-level targets. Across Qwen3-1.7B, 4B, and 8B on five competition-level mathematics benchmarks, \ours{} outperforms strong baselines and sustains its gains through training. These results support the selected subspace as an effective operational proxy for privileged corrections in our setting. We hope our findings motivate further study of which privileged signals
are worth internalizing into a student, and of principled ways to identify
the components of those signals that recur across problems.

\section*{AI use statement}
We used generative AI tools solely to assist with translation, information searches, auxiliary code editing, and language polishing to improve the readability of the manuscript. All AI-assisted outputs were carefully reviewed and verified by the authors. Generative AI was not used to develop the research ideas, methodology, experiments, analyses, interpretations, scientific claims, or conclusions. The authors conducted all substantive research work and take full responsibility for the final content of this paper.

\bibliography{iclr2027_conference}
\bibliographystyle{iclr2027_conference}

\clearpage
\appendix

\section{Limitations}
\label{app:limitations}
We discuss several limitations of \ours{} that we leave to future work:
\begin{itemize}[leftmargin=*, itemsep=2pt, topsep=2pt]
\item First, all models are trained with thinking mode disabled, as in
our main evaluations. Appendix~\ref{app:nothink-think} additionally evaluates
Qwen3-1.7B with thinking mode enabled at inference. We have not
studied training or distillation under thinking mode, and leave
that setting for future work.

\item Second, \ours{} encourages the model to internalize a more complete reasoning process rather than shortcut to the answer, which tends to make its outputs longer and, correspondingly, increases training cost relative to methods that do not. This is a trade-off we accept for more reliable reasoning, and reducing this overhead while retaining the benefit is an interesting direction.

\item Third, our representation matching is applied at the last layer only. Whether matching across all layers, or dynamically selecting which layers to match during training, offers additional benefit remains open, and we regard it as a promising direction for future exploration.
\end{itemize}

\section{What the Leading Low-Rank Subspace Captures}
\label{app:svd-discussion}

\subsection{Interpretation and Limitations}
\label{app:svd-theory}

The leading subspace of a pooled shift matrix captures the directions
with the greatest observed energy. This property alone does not identify
which directions will improve downstream learning. Answer-specific
information could occupy a leading direction, while a useful but
infrequent reasoning pattern could lie in the spectral tail. We therefore
treat the leading subspace as an empirical proxy for cross-problem
structure, not as a decomposition of transferable and non-transferable
knowledge. Three design choices motivate its use in our setting:

\begin{itemize}[leftmargin=*, itemsep=2pt, topsep=2pt]
    \item \textbf{Source-side abstraction.} We form the shifts from compact
    hints rather than full solutions. This removes much of the
    solution-specific detail before estimating the subspace, although it
    does not guarantee that the remaining shift contains only useful
    guidance.

    \item \textbf{Cross-problem checks.} We examine whether the selected
    directions recur among related problems and how their geometry
    relates to downstream performance. The overlap measure in
    Section~\ref{app:subspace-probe} does not use singular values after
    the subspaces have been estimated. Recurrence provides an empirical
    check on the selected directions, but does not by itself establish
    their usefulness.

    \item \textbf{Cluster-specific estimation.} We estimate a separate
    subspace within each content cluster. A pattern that is weak in a
    global pool may become more prominent among related problems.
    This is a motivation for routing rather than a guarantee that every
    useful low-frequency pattern will be retained. Removing routing
    reduces average accuracy on both backbones
    (Table~\ref{tab:ablation}).
\end{itemize}

The following analyses test what the selected subspace captures and
where this interpretation has limits. In particular, concentrated and
consistent shifts need not be useful when the privileged context
provides little actionable guidance
(Appendix~\ref{app:concept}).

\subsection{What the Retained Subspace Captures}
\label{app:subspace-probe}

\paragraph{Setting.}
The leading subspace might reflect properties of the privileged context
that are unrelated to useful guidance. We therefore compare correct
hints with two controls and examine both subspace geometry and
downstream learning. A \emph{wrong hint} matches the correct hint in style and length but
states an incorrect answer. A \emph{cross-problem hint} comes from a
different problem. Both controls are matched to the correct hint in
length and shift norm $\lVert\Delta\rVert$. These controls test whether
the observed differences persist when those two properties are held
fixed. We estimate every subspace using the base model before
distillation, then measure its cross-problem overlap and its alignment
with the correct-hint subspace. The geometric probe uses 48 OpenR1
problems across four clusters of 12, with overlap measured within
clusters. Downstream results use Qwen3-8B trained on OpenR1. For two problems with estimated subspaces spanned by
$V_i,V_j\in\mathbb{R}^{d\times r}$, we define their overlap as
\begin{equation}
\label{eq:overlap}
\mathrm{overlap}(i,j)
=
\frac{1}{r}\lVert V_i^{\top}V_j\rVert_F^2
=
\frac{1}{r}\sum_{k=1}^{r}\cos^2\theta_k,
\end{equation}
where $\theta_k$ are the principal angles between the subspaces. This
quantity lies in $[0,1]$ and depends on the spanned directions rather
than their singular values. We average it over problem pairs within
each cluster. For randomly oriented $r$-dimensional subspaces, the
expected overlap is $r/d$. Here $r=128$ and $d=4096$, giving a reference
value of approximately $0.031$.

\begin{figure}[h]
\centering
\includegraphics[width=\linewidth]{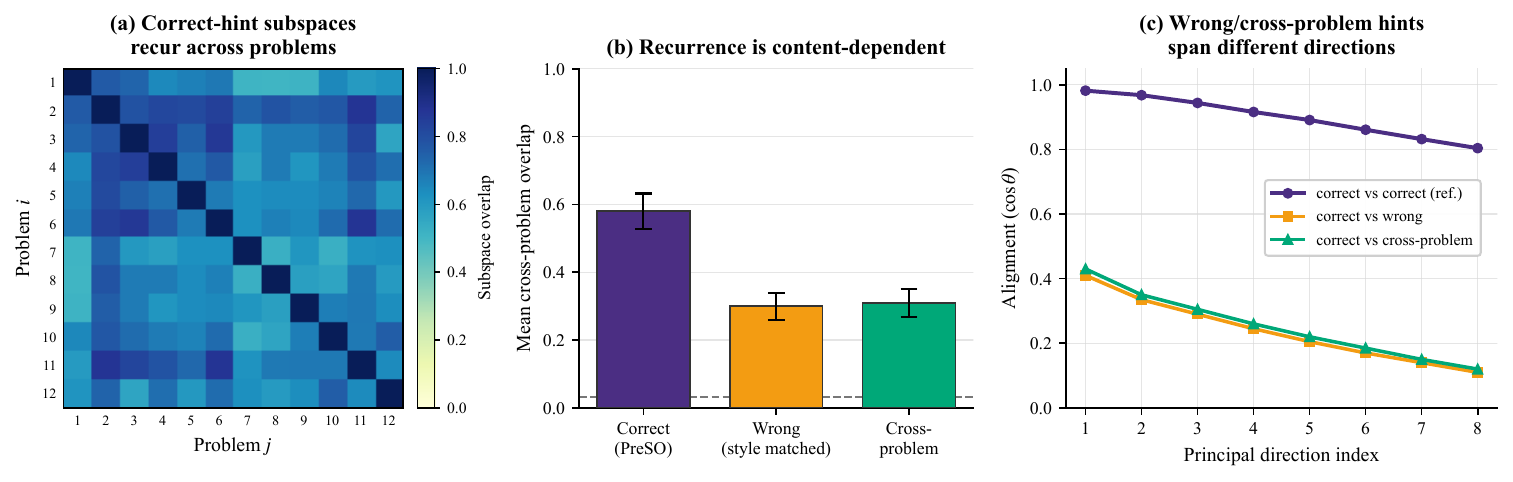}
\vspace{-12pt}
\caption{\textbf{Correct hints induce more consistent subspaces across
related problems.} All panels use Qwen3-8B on OpenR1, with subspaces
estimated on the base model before distillation. (a) Pairwise overlap
among correct-hint subspaces within a cluster. (b) Mean within-cluster
overlap for each condition. The dashed line marks the random-subspace
reference $r/d\approx0.031$. (c) Alignment of the leading
correct-hint directions with an independent correct-hint reference
and with each control.}
\label{fig:subspace-probe}
\end{figure}

\begin{figure}[h]
\centering
\vspace{-12pt}
\includegraphics[width=\linewidth]{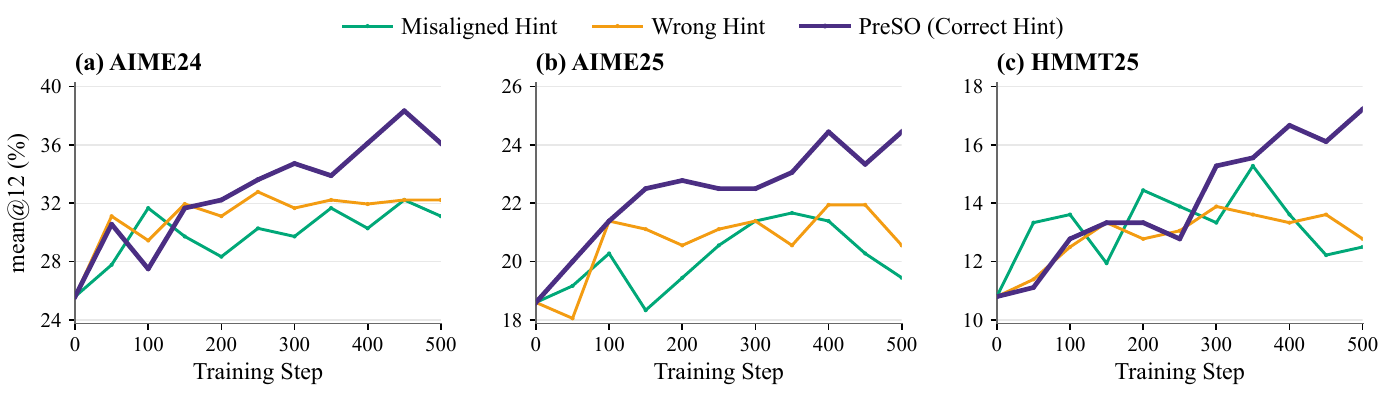}
\vspace{-12pt}
\caption{\textbf{Correct hints yield more sustained downstream gains.}
Accuracy (mean@12) during training on AIME24, AIME25, and HMMT25 for
Qwen3-8B trained on OpenR1 with correct, wrong, or cross-problem hints.}
\label{fig:content-downstream}
\end{figure}

\paragraph{Correct-hint subspaces show stronger recurrence.}
Figure~\ref{fig:subspace-probe}(a) shows substantial overlap among
correct-hint subspaces from different problems within a cluster.
Their mean overlap is approximately $0.58$, compared with $0.30$ for
wrong hints and $0.31$ for cross-problem hints
(Figure~\ref{fig:subspace-probe}b). Both controls remain above the
random-subspace reference, so recurrence is not unique to correct
hints. Nevertheless, correct hints show substantially greater
overlap despite the matched length and shift norm. These results
indicate that the content of the privileged context affects the
geometry beyond those matched properties.

\paragraph{The controls favor different directions.}
Figure~\ref{fig:subspace-probe}(c) shows that independently estimated
correct-hint subspaces remain aligned above $0.8$ across the leading
directions. In contrast, alignment with either control falls from about
$0.42$ on the first direction to $0.11$ by the eighth. The controls
thus alter the prominent directions rather than merely weakening the
correct-hint shift.

\paragraph{Geometry and downstream learning are associated.}
With correct hints, accuracy improves and remains higher through the
end of training on the three benchmarks in
Figure~\ref{fig:content-downstream}. Wrong and cross-problem hints
produce smaller early gains, then plateau or decline. In this
controlled comparison, the condition with greater subspace overlap
and alignment also yields more sustained downstream gains. This
association supports using the leading subspace as a candidate source
of guidance, while the concept-only result in
Appendix~\ref{app:concept} shows why geometry alone is insufficient
to establish that guidance is useful.

\subsection{Direction versus Correction Norm}
\label{app:energy-matched}

\begin{figure}[h]
    \centering
    \vspace{-5pt}
    \includegraphics[width=\linewidth]{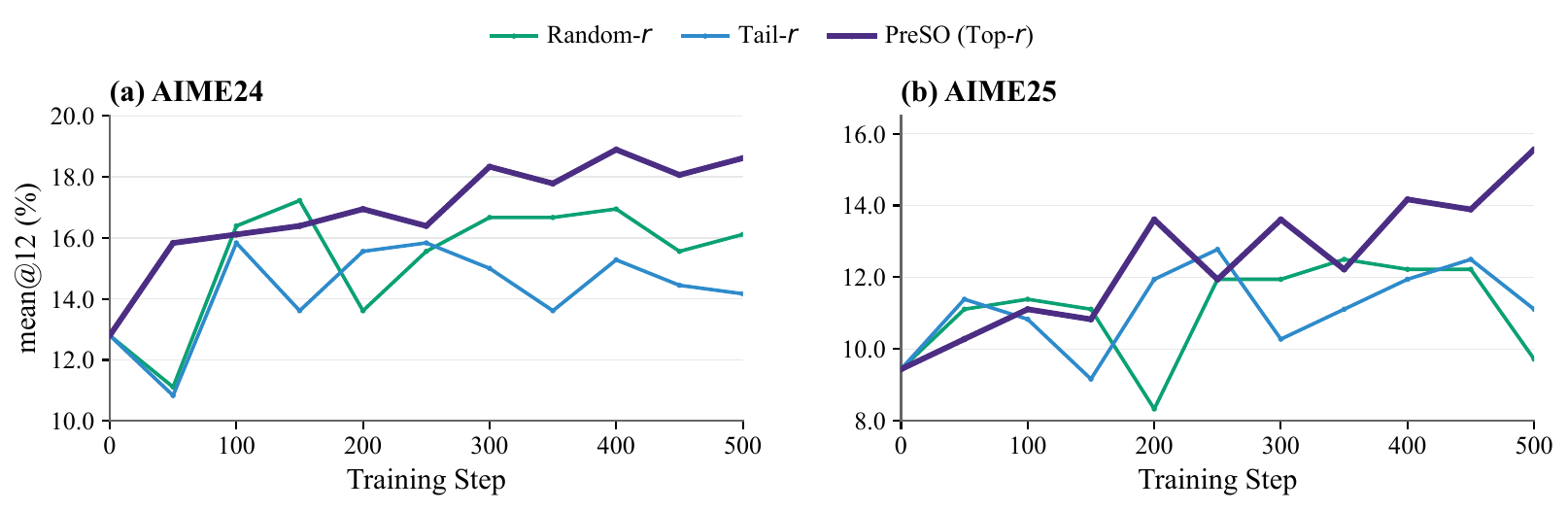}
    \caption{\textbf{Top-$r$ remains ahead after matching correction
    norms.} Accuracy (mean@12) during training on AIME24 and AIME25
    with Qwen3-1.7B. Random-$r$ and Tail-$r$ corrections are rescaled
    to match the pre-normalization norm $\lVert P\Delta\rVert$ of
    Top-$r$.}
    \label{fig:energy-matched}
\end{figure}

Top-$r$, Random-$r$, and Tail-$r$ retain different norms
$\lVert P\Delta\rVert$. To test whether this difference alone
accounts for the subspace-ablation results, we rescale the Random-$r$
and Tail-$r$ corrections to match the norm of the Top-$r$ correction.
This equalizes the correction norm before target construction, while
leaving the selected directions different. Figure~\ref{fig:energy-matched} shows that Top-$r$ remains ahead on
both benchmarks, whereas the rescaled alternatives plateau lower or
decline late in training. The advantage therefore cannot be
explained solely by the retained correction norm. Because the target
is subsequently normalized and the loss is evaluated over the full
hidden vector, this intervention does not equate the complete
training signals across variants. It nevertheless provides additional
evidence that the choice of directions matters.

\subsection{Effect of the Subspace Rank}
\label{app:rank-ablation}

\begin{figure}[t]
    \centering
    \includegraphics[width=\linewidth]{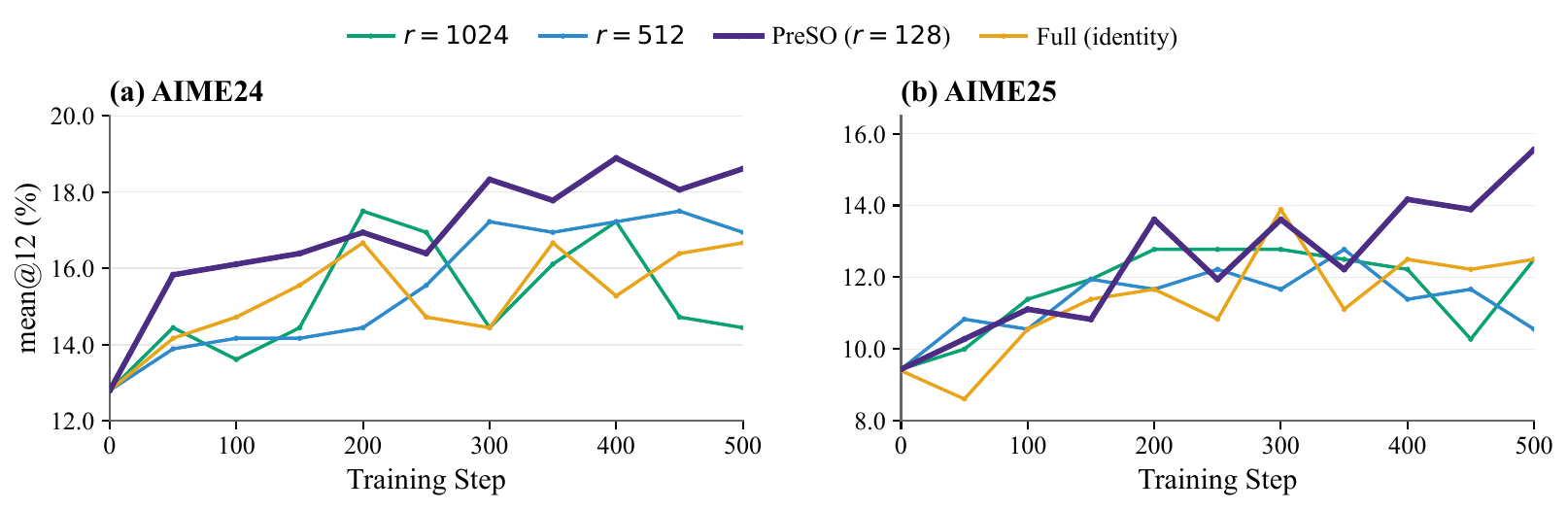}
    \caption{\textbf{A compact leading subspace performs best in this
    rank sweep.} Accuracy (mean@12) during training on AIME24 and
    AIME25 with Qwen3-1.7B. The variants retain different numbers
    of leading directions from the same pooled shift, with Full
    applying no projection.}
    \label{fig:rank-ablation}
    \vspace{-5pt}
\end{figure}

The main-text sweep selects $r$ within a narrow range. Here we
compare $r\in\{128,512,1024\}$ with the full-shift variant
($P=I$) on Qwen3-1.7B. All variants retain the leading directions
of the same pooled shift, with only the number of retained directions
changed. Figure~\ref{fig:rank-ablation} shows that $r=128$ achieves the
strongest and most stable performance in this comparison, while
larger ranks and the full-shift variant plateau or fluctuate more.
As $r$ increases, the target incorporates additional, lower-energy
directions. The fraction of observed shift energy retained by
Top-$r$ is determined by the leading singular values, not by
$r/d$. This experiment does not establish that the added directions
are answer-specific or intrinsically harmful. It shows that
retaining more of the observed shift does not improve learning in
this setting. Because rank also changes the correction norm, this
sweep does not isolate dimensionality from correction magnitude.

\subsection{Summary}

The controlled hint comparison shows that correct hints induce
leading subspaces with greater cross-problem overlap and alignment,
alongside more sustained downstream gains. Norm matching shows that
the advantage of Top-$r$ cannot be explained solely by the size of
the retained correction. The rank sweep shows that retaining more
directions is not necessarily better. Together, these observations
support the leading subspace as an effective empirical choice for
constructing privileged corrections in our setting. They do not
establish that SVD separates useful from useless knowledge, or that
low rank and recurrence alone guarantee downstream benefit.

\section{Supplementary Experiments}
\label{app:more_experiments}
\subsection{Full Experimental Results on DeepMath}
\label{app:last-ckpt}

\begin{figure}[h]
\centering
\includegraphics[width=\linewidth]{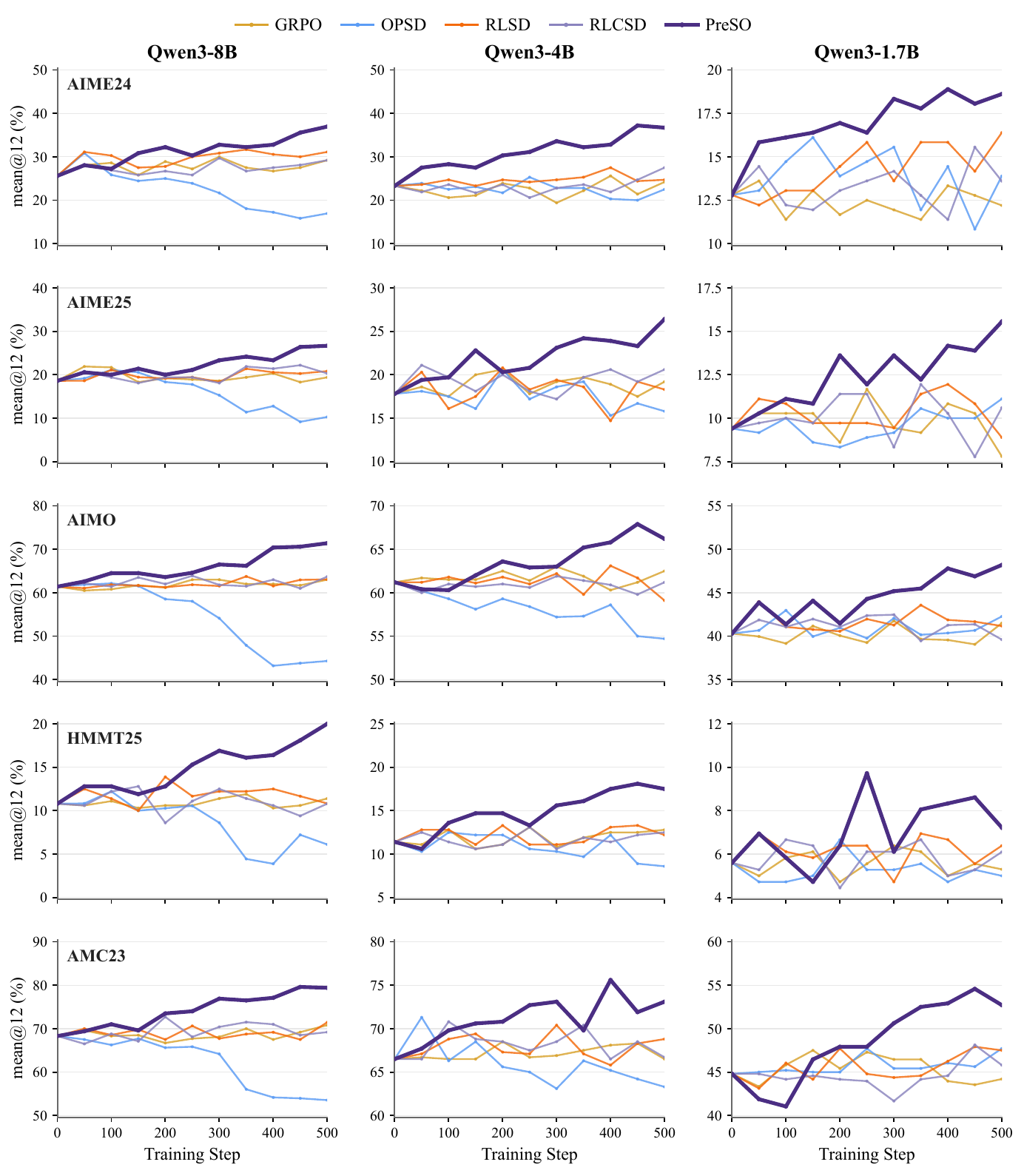}
\vspace{-12pt}
\caption{\textbf{Full training curves on DeepMath-103K.}
mean@12 over training steps for \ours{} and all baselines, across Qwen3-8B, 4B, and 1.7B (columns) on the five competition-level benchmarks.}
\label{fig:full-curves}
\end{figure}

The main tables report the best checkpoint for every method. To show that our gains
do not depend on checkpoint selection, Table~\ref{tb:last_ckpt} reports mean@12 at
the last checkpoint, the model after the full training run, under the same setting.
\ours{} remains the best method on every benchmark and every base model, reaching
$46.9$, $44.0$, and $28.6$ on average for Qwen3-8B, 4B, and 1.7B, exceeding the
strongest baseline by $7.4$, $6.3$, and $4.5$ points. The gap widens relative to the
best-checkpoint tables, because the baselines are far more sensitive to when training
stops. This is clearest for OPSD, the pure distillation baseline, which collapses by
the last checkpoint and falls $10.7$ points below the base model on Qwen3-8B, whereas
its best-checkpoint result is competitive. GRPO and the hybrid baselines RLSD and
RLCSD hold up better but still trail \ours{} at every scale. The full training curves
in Figure~\ref{fig:full-curves} tell the same story across all steps: \ours{} rises
steadily and stays on top of every baseline throughout training on all three model
sizes and all five benchmarks. These results show that \ours{} achieves higher final-checkpoint
accuracy and more stable training dynamics across the evaluated
settings, consistent with the main-text findings.

\begin{table*}[h]
\caption{Experimental results (\%) on mathematical reasoning benchmarks. For every method we
report mean@12 at the last checkpoint. Under
each base model, the best and second-best results are marked in \textbf{bold} and \underline{underlined}, respectively.}
\vspace{0pt}
\centering
\small
\renewcommand{\arraystretch}{1.20}
\setlength{\tabcolsep}{5pt}

\begin{tabularx}{0.96\textwidth}{l|*{6}{Y}}
\toprule
\textbf{Method}
& \textbf{AIME24}
& \textbf{AIME25}
& \textbf{AIMO}
& \textbf{HMMT25}
& \textbf{AMC23}
& \textbf{Avg.} \\
\midrule

\textbf{Qwen3-8B}
& 25.6 \basex{0.0}
& 18.6 \basex{0.0}
& 61.4 \basex{0.0}
& 10.8 \basex{0.0}
& 68.3 \basex{0.0}
& 36.9 \basex{0.0} \\

\hspace*{1em}+ GRPO
& 29.2 \uprise{3.6}
& 19.4 \uprise{0.8}
& 63.0 \uprise{1.6}
& \underline{11.4} \uprise{0.6}
& 70.8 \uprise{2.5}
& 38.8 \uprise{1.9} \\

\hspace*{1em}+ OPSD
& 16.9 \down{8.7}
& 10.3 \down{8.3}
& 44.3 \down{17.1}
& 6.1 \down{4.7}
& 53.5 \down{14.8}
& 26.2 \down{10.7} \\

\hspace*{1em}+ RLSD
& \underline{31.1} \uprise{5.5}
& \underline{20.8} \uprise{2.2}
& 63.1 \uprise{1.7}
& 10.8 \basex{0.0}
& \underline{71.5} \uprise{3.2}
& \underline{39.5} \uprise{2.6} \\

\hspace*{1em}+ RLCSD
& 29.2 \uprise{3.6}
& 20.3 \uprise{1.7}
& \underline{63.7} \uprise{2.3}
& 10.8 \basex{0.0}
& 69.2 \uprise{0.9}
& 38.6 \uprise{1.7} \\

\rowcolor{blue!10}
\hspace*{1em}+ \textbf{\ours{}}
& \textbf{37.0} \uprise{11.4}
& \textbf{26.7} \uprise{8.1}
& \textbf{71.4} \uprise{10.0}
& \textbf{20.0} \uprise{9.2}
& \textbf{79.4} \uprise{11.1}
& \textbf{46.9} \uprise{10.0} \\

\midrule

\textbf{Qwen3-4B}
& 23.3 \basex{0.0}
& 17.8 \basex{0.0}
& 61.2 \basex{0.0}
& 11.4 \basex{0.0}
& 66.5 \basex{0.0}
& 36.0 \basex{0.0} \\

\hspace*{1em}+ GRPO
& 24.2 \uprise{0.9}
& 19.2 \uprise{1.4}
& \underline{62.5} \uprise{1.3}
& \underline{12.8} \uprise{1.4}
& 66.5 \basex{0.0}
& 37.0 \uprise{1.0} \\

\hspace*{1em}+ OPSD
& 22.5 \down{0.8}
& 15.8 \down{2.0}
& 54.7 \down{6.5}
& 8.6 \down{2.8}
& 63.3 \down{3.2}
& 33.0 \down{3.0} \\

\hspace*{1em}+ RLSD
& 24.7 \uprise{1.4}
& 18.3 \uprise{0.5}
& 59.1 \down{2.1}
& 12.2 \uprise{0.8}
& \underline{68.8} \uprise{2.3}
& 36.6 \uprise{0.6} \\

\hspace*{1em}+ RLCSD
& \underline{27.5} \uprise{4.2}
& \underline{20.6} \uprise{2.8}
& 61.2 \basex{0.0}
& 12.5 \uprise{1.1}
& 66.7 \uprise{0.2}
& \underline{37.7} \uprise{1.7} \\

\rowcolor{blue!10}
\hspace*{1em}+ \textbf{PreSO}
& \textbf{36.7} \uprise{13.4}
& \textbf{26.4} \uprise{8.6}
& \textbf{66.2} \uprise{5.0}
& \textbf{17.5} \uprise{6.1}
& \textbf{73.1} \uprise{6.6}
& \textbf{44.0} \uprise{8.0} \\

\midrule

\textbf{Qwen3-1.7B}
& 12.8 \basex{0.0}
& 9.4 \basex{0.0}
& 40.3 \basex{0.0}
& 5.6 \basex{0.0}
& 44.8 \basex{0.0}
& 22.6 \basex{0.0} \\

\hspace*{1em}+ GRPO
& 13.1 \uprise{0.3}
& \underline{11.1} \uprise{1.7}
& 41.5 \uprise{1.2}
& 6.1 \uprise{0.5}
& 46.7 \uprise{1.9}
& 23.7 \uprise{1.1} \\

\hspace*{1em}+ OPSD
& 13.9 \uprise{1.1}
& \underline{11.1} \uprise{1.7}
& \underline{42.3} \uprise{2.0}
& 5.0 \down{0.6}
& \underline{47.7} \uprise{2.9}
& 24.0 \uprise{1.4} \\

\hspace*{1em}+ RLSD
& \underline{16.4} \uprise{3.6}
& 8.9 \down{0.5}
& 41.2 \uprise{0.9}
& \underline{6.4} \uprise{0.8}
& 47.5 \uprise{2.7}
& \underline{24.1} \uprise{1.5} \\

\hspace*{1em}+ RLCSD
& 13.6 \uprise{0.8}
& 10.6 \uprise{1.2}
& 42.0 \uprise{1.7}
& 6.1 \uprise{0.5}
& 45.8 \uprise{1.0}
& 23.6 \uprise{1.0} \\

\rowcolor{blue!10}
\hspace*{1em}+ \textbf{PreSO}
& \textbf{18.6} \uprise{5.8}
& \textbf{15.6} \uprise{6.2}
& \textbf{48.2} \uprise{7.9}
& \textbf{8.1} \uprise{2.5}
& \textbf{52.7} \uprise{7.9}
& \textbf{28.6} \uprise{6.0} \\

\bottomrule
\end{tabularx}
\label{tb:last_ckpt}
\end{table*}

\subsection{Baselines with Full-Solution Privileged Information}
\label{app:solution-pi}

In the main results, OPSD and RLSD receive the same GPT-5.5 hints as
\ours{}, ensuring that the methods use identical privileged information.
Here we instead provide these baselines with full reference solutions,
as in their original settings, while keeping all other hyperparameters
fixed. Table~\ref{tb:solution-pi} reports results at the best and final
checkpoints. The two sources yield similar peak performance, but
solution-based OPSD declines more sharply thereafter. On Qwen3-8B, its
average falls from $39.0$ to $21.2$ with solutions, compared with
$38.8$ to $26.2$ with hints. On HMMT25, the solution-based score drops
from $11.7$ to $2.8$. The same pattern holds at $4$B and $1.7$B.
RLSD, which also uses outcome feedback, is less sensitive to the
privileged source. Together with the source-side ablation in
Section~\ref{sec:ablation}, these results suggest that the additional
detail in full solutions can make distillation less stable over
training. They support compact hints as a source-side choice, without
attributing the full gain of \ours{} to abstraction alone.

\begin{table*}[!h]
\caption{Comparison of solution-based and hint-based variants of OPSD
and RLSD on mathematical reasoning benchmarks (\%).
Each cell reports the best / last checkpoint results.
Avg. denotes the arithmetic mean across the five benchmarks.}
\vspace{0pt}
\centering
\small
\renewcommand{\arraystretch}{1.20}
\setlength{\tabcolsep}{4pt}

\begin{tabularx}{\textwidth}{ll|*{6}{Y}}
\toprule
\textbf{Method}
& \textbf{Variant}
& \textbf{AIME24}
& \textbf{AIME25}
& \textbf{AIMO}
& \textbf{HMMT25}
& \textbf{AMC23}
& \textbf{Avg.} \\
\midrule

\multicolumn{8}{l}{\textbf{Qwen3-8B}} \\
\midrule

OPSD & Solution
& 30.3 / 10.6
& 20.8 / 7.5
& 61.7 / 38.4
& 11.7 / 2.8
& 70.4 / 46.5
& 39.0 / 21.2 \\

OPSD & Hint
& 30.8 / 16.9
& 21.1 / 10.3
& 62.1 / 44.3
& 12.2 / 6.1
& 67.7 / 53.5
& 38.8 / 26.2 \\

\addlinespace[3pt]

RLSD & Solution
& 29.7 / 25.3
& 23.3 / 20.8
& 63.7 / 62.3
& 12.5 / 11.4
& 70.6 / 69.4
& 40.0 / 37.8 \\

RLSD & Hint
& 31.7 / 31.1
& 21.4 / 20.8
& 63.8 / 63.1
& 13.9 / 10.8
& 71.5 / 71.5
& 40.5 / 39.5 \\

\midrule
\multicolumn{8}{l}{\textbf{Qwen3-4B}} \\
\midrule

OPSD & Solution
& 25.0 / 17.5
& 20.6 / 13.6
& 60.0 / 51.3
& 11.9 / 10.3
& 68.3 / 60.6
& 37.2 / 30.7 \\

OPSD & Hint
& 25.3 / 22.5
& 20.6 / 15.8
& 60.2 / 54.7
& 12.5 / 8.6
& 71.3 / 63.3
& 38.0 / 33.0 \\

\addlinespace[3pt]

RLSD & Solution
& 24.7 / 22.5
& 20.8 / 18.6
& 60.9 / 60.7
& 13.3 / 13.3
& 69.6 / 68.3
& 37.9 / 36.7 \\

RLSD & Hint
& 27.5 / 24.7
& 20.8 / 18.3
& 63.1 / 59.1
& 13.3 / 12.2
& 70.4 / 68.8
& 39.0 / 36.6 \\

\midrule
\multicolumn{8}{l}{\textbf{Qwen3-1.7B}} \\
\midrule

OPSD & Solution
& 15.0 / 10.6
& 11.7 / 7.2
& 42.2 / 40.2
& 5.8 / 5.3
& 46.7 / 42.3
& 24.3 / 21.1 \\

OPSD & Hint
& 16.1 / 13.9
& 11.1 / 11.1
& 43.0 / 42.3
& 6.7 / 5.0
& 47.7 / 47.7
& 24.9 / 24.0 \\

\addlinespace[3pt]

RLSD & Solution
& 14.7 / 11.9
& 11.1 / 10.6
& 43.1 / 41.7
& 6.7 / 6.4
& 47.5 / 47.5
& 24.6 / 23.6 \\

RLSD & Hint
& 16.4 / 16.4
& 11.9 / 8.9
& 44.0 / 41.2
& 6.9 / 6.4
& 47.9 / 47.5
& 25.4 / 24.1 \\

\bottomrule
\end{tabularx}
\label{tb:solution-pi}
\end{table*}

\subsection{Experimental Results on Open-R1}
\label{app:openr1_result}
To verify that the effectiveness of \ours{} does not hinge on a particular training corpus, we repeat the full comparison with Open-R1~\citep{luffy} in place of DeepMath-103K, keeping all other settings unchanged. Tables~\ref{tb:openr1_best} and~\ref{tb:openr1_last} report the best-checkpoint and last-checkpoint results, respectively. The overall picture matches our main experiments. At the best checkpoint, \ours{} attains the highest average accuracy on all three backbones, improving over the base model by $\mathbf{7.0}$, $\mathbf{8.7}$, and $\mathbf{7.3}$ points on Qwen3-1.7B, 4B, and 8B, and ranking first on nearly every individual benchmark. The last-checkpoint results further confirm its stability: \ours{} stays essentially unchanged from its best checkpoint at every scale, whereas OPSD degrades sharply once trained past its peak, falling below the base model on Qwen3-8B and Qwen3-4B.

\begin{table*}[!t]
\caption{Experimental results (\%) on mathematical reasoning benchmarks with models trained on Open-R1. For each method, we report the highest mean@12 across evaluated checkpoints and their average (Avg.). The best and second-best scores are highlighted in \textbf{bold} and \underline{underlined}, respectively.}
\centering
\renewcommand{\arraystretch}{1.25}

\resizebox{\textwidth}{!}{
\begin{tabular}{l|ccccc|c}
\toprule
\textbf{Method}
& \textbf{AIME24}
& \textbf{AIME25}
& \textbf{AIMO}
& \textbf{HMMT25}
& \textbf{AMC23}
& \textbf{Avg.} \\
\midrule

\textbf{Qwen3-8B}
& 25.6\basex{0.0}
& 18.6\basex{0.0}
& 61.4\basex{0.0}
& 10.8\basex{0.0}
& 68.3\basex{0.0}
& 36.9\basex{0.0} \\

\hspace*{1em}+ GRPO
& 30.6\uprise{5.0}
& 23.6\uprise{5.0}
& 63.7\uprise{2.3}
& 14.1\uprise{3.3}
& 71.3\uprise{3.0}
& 40.7\uprise{3.8} \\

\hspace*{1em}+ OPSD
& 31.9\uprise{6.3}
& 21.9\uprise{3.3}
& 62.1\uprise{0.7}
& 12.2\uprise{1.4}
& 69.8\uprise{1.5}
& 39.6\uprise{2.7} \\

\hspace*{1em}+ RLSD
& 30.6\uprise{5.0}
& \underline{23.7}\uprise{5.1}
& 63.7\uprise{2.3}
& \underline{14.2}\uprise{3.4}
& \underline{71.7}\uprise{3.4}
& \underline{40.8}\uprise{3.9} \\

\hspace*{1em}+ RLCSD
& \underline{33.1}\uprise{7.5}
& 23.3\uprise{4.7}
& \underline{64.0}\uprise{2.6}
& 12.8\uprise{2.0}
& 71.0\uprise{2.7}
& \underline{40.8}\uprise{3.9} \\

\rowcolor{blue!10}
\hspace*{1em}+ \textbf{\ours{}}
& \textbf{37.0} \uprise{11.4}
& \textbf{24.4} \uprise{5.8}
& \textbf{67.2} \uprise{5.8}
& \textbf{17.2} \uprise{6.4}
& \textbf{75.0} \uprise{6.7}
& \textbf{44.2} \uprise{7.3} \\

\midrule

\textbf{Qwen3-4B}
& 23.3\basex{0.0}
& 17.8\basex{0.0}
& 61.2\basex{0.0}
& 11.4\basex{0.0}
& 66.5\basex{0.0}
& 36.0\basex{0.0} \\

\hspace*{1em}+ GRPO
& \underline{25.6}\uprise{2.3}
& \underline{21.7}\uprise{3.9}
& \underline{62.3}\uprise{1.1}
& 13.3\uprise{1.9}
& 70.2\uprise{3.7}
& 38.6\uprise{2.6} \\

\hspace*{1em}+ OPSD
& 25.3\uprise{2.0}
& 19.2\uprise{1.4}
& 60.4\down{0.8}
& \underline{14.4}\uprise{3.0}
& 69.0\uprise{2.5}
& 37.7\uprise{1.7} \\

\hspace*{1em}+ RLSD
& \underline{25.6}\uprise{2.3}
& \underline{21.7}\uprise{3.9}
& 62.2\uprise{1.0}
& 13.3\uprise{1.9}
& \underline{70.6}\uprise{4.1}
& \underline{38.7}\uprise{2.7} \\

\hspace*{1em}+ RLCSD
& 25.0\uprise{1.7}
& 21.1\uprise{3.3}
& 60.4\down{0.8}
& 13.1\uprise{1.7}
& 69.4\uprise{2.9}
& 37.8\uprise{1.8} \\

\rowcolor{blue!10}
\hspace*{1em}+ \textbf{PreSO}
& \textbf{34.4}\uprise{11.1}
& \textbf{24.4}\uprise{6.6}
& \textbf{68.2}\uprise{7.0}
& \textbf{19.2}\uprise{7.8}
& \textbf{77.5}\uprise{11.0}
& \textbf{44.7}\uprise{8.7} \\

\midrule

\textbf{Qwen3-1.7B}
& 12.8\basex{0.0}
& 9.4\basex{0.0}
& 40.3\basex{0.0}
& 5.6\basex{0.0}
& 44.8\basex{0.0}
& 22.6\basex{0.0} \\

\hspace*{1em}+ GRPO
& 15.6\uprise{2.8}
& 11.4\uprise{2.0}
& 41.9\uprise{1.6}
& \underline{6.9}\uprise{1.3}
& \underline{48.5}\uprise{3.7}
& 24.9\uprise{2.3} \\

\hspace*{1em}+ OPSD
& 16.4\uprise{3.6}
& \underline{13.6}\uprise{4.2}
& 42.6\uprise{2.3}
& 6.7\uprise{1.1}
& 47.1\uprise{2.3}
& 25.3\uprise{2.7} \\

\hspace*{1em}+ RLSD
& \underline{17.5}\uprise{4.7}
& 11.9\uprise{2.5}
& 42.7\uprise{2.4}
& \underline{6.9}\uprise{1.3}
& \underline{48.5}\uprise{3.7}
& \underline{25.5}\uprise{2.9} \\

\hspace*{1em}+ RLCSD
& 15.8\uprise{3.0}
& 11.7\uprise{2.3}
& \underline{43.0}\uprise{2.7}
& \underline{6.9}\uprise{1.3}
& \underline{48.5}\uprise{3.7}
& 25.2\uprise{2.6} \\

\rowcolor{blue!10}
\hspace*{1em}+ \textbf{PreSO}
& \textbf{20.8}\uprise{8.0}
& \textbf{15.6}\uprise{6.2}
& \textbf{48.0}\uprise{7.7}
& \textbf{8.6}\uprise{3.0}
& \textbf{55.2}\uprise{10.4}
& \textbf{29.6}\uprise{7.0} \\

\bottomrule
\end{tabular}
}
\label{tb:openr1_best}
\end{table*}

\begin{table*}[!t]
\caption{Last-checkpoint results (\%) on mathematical reasoning benchmarks with models trained on Open-R1. For each method, we report the mean@12 at the final checkpoint and their average (Avg.). The best and second-best scores are highlighted in \textbf{bold} and \underline{underlined}, respectively.}
\centering
\renewcommand{\arraystretch}{1.25}

\resizebox{\textwidth}{!}{
\begin{tabular}{l|ccccc|c}
\toprule
\textbf{Method}
& \textbf{AIME24}
& \textbf{AIME25}
& \textbf{AIMO}
& \textbf{HMMT25}
& \textbf{AMC23}
& \textbf{Avg.} \\
\midrule

\textbf{Qwen3-8B}
& 25.6\basex{0.0}
& 18.6\basex{0.0}
& 61.4\basex{0.0}
& 10.8\basex{0.0}
& 68.3\basex{0.0}
& 36.9\basex{0.0} \\

\hspace*{1em}+ GRPO
& 30.0\uprise{4.4}
& 20.1\uprise{1.5}
& 63.3\uprise{1.9}
& \underline{14.1}\uprise{3.3}
& 70.0\uprise{1.7}
& 39.5\uprise{2.6} \\

\hspace*{1em}+ OPSD
& 15.6\down{10.0}
& 11.1\down{7.5}
& 42.3\down{19.1}
& 6.9\down{3.9}
& 66.5\down{1.8}
& 28.5\down{8.4} \\

\hspace*{1em}+ RLSD
& 30.3\uprise{4.7}
& 21.7\uprise{3.1}
& 63.5\uprise{2.1}
& 11.8\uprise{1.0}
& \underline{71.0}\uprise{2.7}
& 39.7\uprise{2.8} \\

\hspace*{1em}+ RLCSD
& \underline{33.1}\uprise{7.5}
& \underline{23.3}\uprise{4.7}
& \underline{64.0}\uprise{2.6}
& 12.0\uprise{1.2}
& \underline{71.0}\uprise{2.7}
& \underline{40.7}\uprise{3.8} \\

\rowcolor{blue!10}
\hspace*{1em}+ \textbf{\ours{}}
& \textbf{37.0} \uprise{11.4}
& \textbf{24.4} \uprise{5.8}
& \textbf{67.2} \uprise{5.8}
& \textbf{17.2} \uprise{6.4}
& \textbf{75.0} \uprise{6.7}
& \textbf{44.2} \uprise{7.3} \\

\midrule

\textbf{Qwen3-4B}
& 23.3\basex{0.0}
& 17.8\basex{0.0}
& \underline{61.2}\basex{0.0}
& 11.4\basex{0.0}
& 66.5\basex{0.0}
& 36.0\basex{0.0} \\

\hspace*{1em}+ GRPO
& \underline{25.0}\uprise{1.7}
& 20.3\uprise{2.5}
& 60.9\down{0.3}
& \underline{13.3}\uprise{1.9}
& \underline{70.2}\uprise{3.7}
& \underline{37.9}\uprise{1.9} \\

\hspace*{1em}+ OPSD
& 19.4\down{3.9}
& 16.4\down{1.4}
& 55.0\down{6.2}
& 10.0\down{1.4}
& 64.4\down{2.1}
& 33.0\down{3.0} \\

\hspace*{1em}+ RLSD
& 23.9\uprise{0.6}
& 18.9\uprise{1.1}
& 60.1\down{1.1}
& 11.1\down{0.3}
& \underline{70.2}\uprise{3.7}
& 36.8\uprise{0.8} \\

\hspace*{1em}+ RLCSD
& 23.9\uprise{0.6}
& \underline{21.1}\uprise{3.3}
& 60.4\down{0.8}
& 13.1\uprise{1.7}
& 69.4\uprise{2.9}
& 37.6\uprise{1.6} \\

\rowcolor{blue!10}
\hspace*{1em}+ \textbf{PreSO}
& \textbf{34.4}\uprise{11.1}
& \textbf{24.4}\uprise{6.6}
& \textbf{68.2}\uprise{7.0}
& \textbf{18.9}\uprise{7.5}
& \textbf{77.5}\uprise{11.0}
& \textbf{44.7}\uprise{8.7} \\

\midrule

\textbf{Qwen3-1.7B}
& 12.8\basex{0.0}
& 9.4\basex{0.0}
& 40.3\basex{0.0}
& 5.6\basex{0.0}
& 44.8\basex{0.0}
& 22.6\basex{0.0} \\

\hspace*{1em}+ GRPO
& 15.6\uprise{2.8}
& 11.4\uprise{2.0}
& 41.9\uprise{1.6}
& \underline{6.9}\uprise{1.3}
& \underline{48.5}\uprise{3.7}
& 24.9\uprise{2.3} \\

\hspace*{1em}+ OPSD
& 13.6\uprise{0.8}
& 11.1\uprise{1.7}
& 40.4\uprise{0.1}
& 5.0\down{0.6}
& 44.2\down{0.6}
& 22.9\uprise{0.3} \\

\hspace*{1em}+ RLSD
& 14.7\uprise{1.9}
& 10.3\uprise{0.9}
& 41.5\uprise{1.2}
& 6.7\uprise{1.1}
& \underline{48.5}\uprise{3.7}
& 24.3\uprise{1.7} \\

\hspace*{1em}+ RLCSD
& \underline{15.8}\uprise{3.0}
& \underline{11.7}\uprise{2.3}
& \underline{43.0}\uprise{2.7}
& \underline{6.9}\uprise{1.3}
& \underline{48.5}\uprise{3.7}
& \underline{25.2}\uprise{2.6} \\

\rowcolor{blue!10}
\hspace*{1em}+ \textbf{PreSO}
& \textbf{20.0}\uprise{7.2}
& \textbf{15.6}\uprise{6.2}
& \textbf{48.0}\uprise{7.7}
& \textbf{8.6}\uprise{3.0}
& \textbf{55.2}\uprise{10.4}
& \textbf{29.5}\uprise{6.9} \\

\bottomrule
\end{tabular}
}
\label{tb:openr1_last}
\end{table*}

\subsection{Self-Generated Hints}
\label{app:self-hint}

In the main experiments, GPT-5.5 annotates the privileged hints offline.
Here we replace them with hints generated on-policy by the student,
keeping all other settings fixed. Table~\ref{tb:self-hint} compares the
two sources at the best and final checkpoints. Self-generated hints
improve \ours{} by $0.9$ points on average at the best checkpoint and
$1.4$ at the final checkpoint, although their relative performance
varies across benchmarks. They may better match the student's current
behavior, but this comparison does not isolate the reason for the
difference. On-policy hint generation also increases training time
from $6$h$18$m to $8$h$20$m, roughly one-third more. Given the modest
accuracy gain and additional cost, we use offline GPT-5.5 hints by
default. The self-generated variant shows that \ours{} does not
require an external hint annotator.

\begin{table*}[h]
\caption{\textbf{Effect of the hint source.} Comparison of \ours{} with offline GPT-5.5 hints and with on-policy self-generated hints on Qwen3-1.7B, reporting mean@12 (\%) at the best and last checkpoints, together with total training time. Per-column best results are in \textbf{bold}. Self-generated hints are slightly more accurate on average but increase training time by about a third.}
\vspace{0pt}
\centering
\small
\renewcommand{\arraystretch}{1.20}
\setlength{\tabcolsep}{3pt}

\begin{tabularx}{\textwidth}{ll|*{6}{Y}|c}
\toprule
\textbf{Method}
& \textbf{Hint source}
& \textbf{AIME24}
& \textbf{AIME25}
& \textbf{AIMO}
& \textbf{HMMT25}
& \textbf{AMC23}
& \textbf{Avg.}
& \shortstack{\textbf{Training}\\\textbf{time}} \\
\midrule

\multicolumn{9}{l}{\textbf{Best checkpoint}} \\
\midrule

Base model
& --
& 12.8
& 9.4
& 40.3
& 5.6
& 44.8
& 22.6
& -- \\

PreSO
& GPT-5.5
& 18.9
& \textbf{15.6}
& 48.2
& 9.7
& \textbf{54.6}
& 29.4
& \textbf{6h 18m} \\

PreSO
& Student (online)
& \textbf{21.7}
& 15.0
& \textbf{50.5}
& \textbf{10.3}
& 54.0
& \textbf{30.3}
& 8h 20m \\

\midrule
\multicolumn{9}{l}{\textbf{Last checkpoint}} \\
\midrule

Base model
& --
& 12.8
& 9.4
& 40.3
& 5.6
& 44.8
& 22.6
& -- \\

PreSO
& GPT-5.5
& 18.6
& \textbf{15.6}
& 48.2
& 8.1
& 52.7
& 28.6
& \textbf{6h 18m} \\

PreSO
& Student (online)
& \textbf{21.4}
& 13.9
& \textbf{50.5}
& \textbf{10.3}
& \textbf{54.0}
& \textbf{30.0}
& 8h 20m \\

\bottomrule
\end{tabularx}
\label{tb:self-hint}
\end{table*}

\subsection{Robustness to Random Seeds}
\label{app:seeds}

\begin{figure}[h]
\centering
\includegraphics[width=\linewidth]{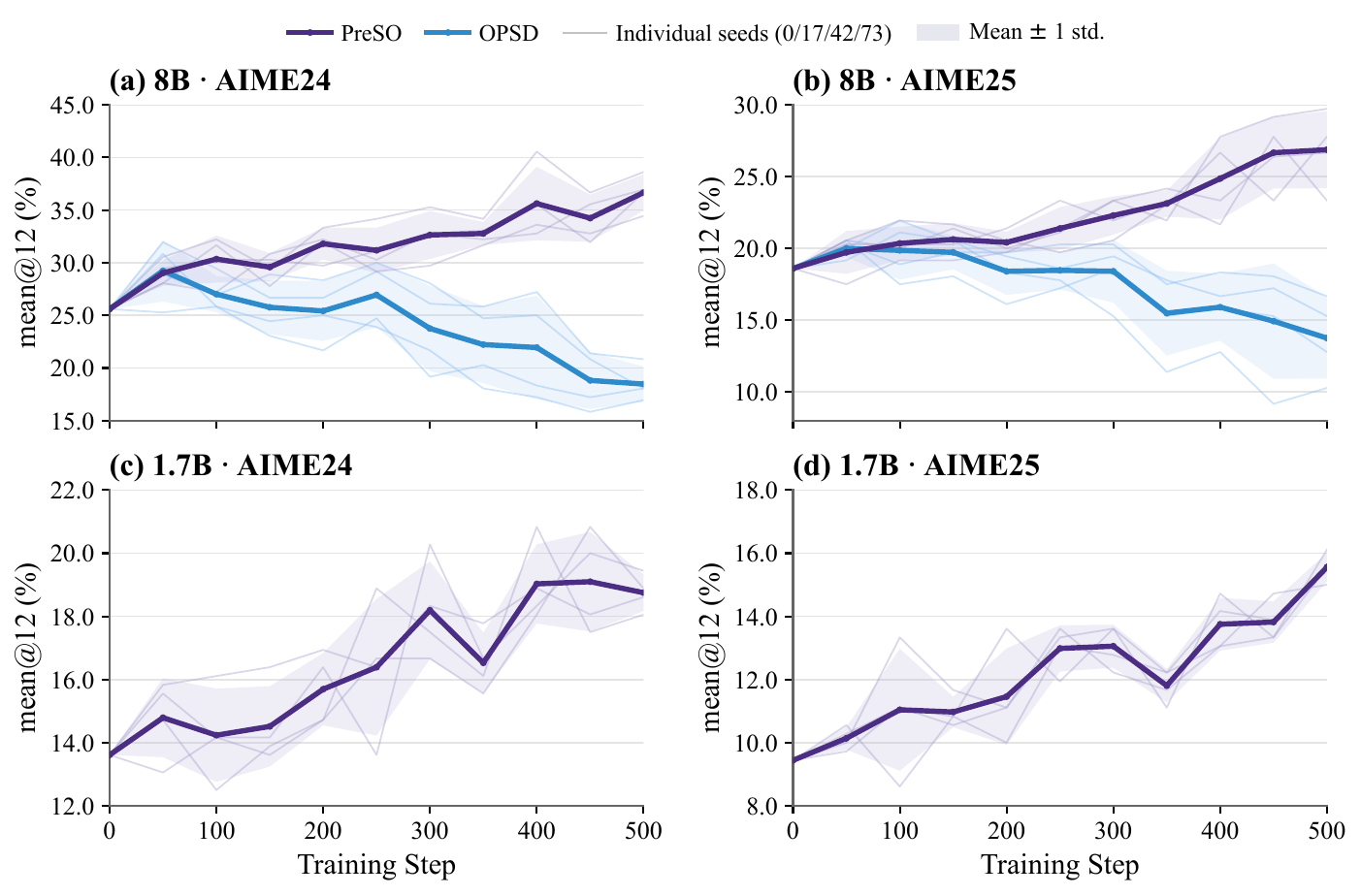}
\vspace{-12pt}
\caption{\textbf{Robustness across random seeds.} Mean@12 during
training with four seeds ($0$, $17$, $42$, and $73$) on AIME24 and
AIME25. We evaluate \ours{} on Qwen3-8B and Qwen3-1.7B, and OPSD
on Qwen3-8B. Thin lines show individual runs, bold lines show
across-seed means, and shaded bands indicate one standard deviation.}
\label{fig:seeds}
\end{figure}

For efficiency, the main results use a single fixed seed. To assess
seed sensitivity, we repeat \ours{} on Qwen3-8B and Qwen3-1.7B with
four seeds ($0$, $17$, $42$, and $73$). We also repeat OPSD on
Qwen3-8B under the same seeds. Figure~\ref{fig:seeds} shows individual
runs and the across-seed mean and standard deviation. On Qwen3-8B,
\ours{} improves across seeds on both AIME24 and AIME25, reaching
approximately $36.7$ and $26.8$ on average at the final checkpoint.
OPSD instead peaks early and declines late under all four seeds on
both benchmarks. \ours{} also shows an overall upward trend across
seeds on Qwen3-1.7B, ending near $19$ on AIME24 and $15.5$ on AIME25.
The main-run trajectory remains close to the across-seed mean. These results reduce the concern that \ours{}'s sustained gains, or its
late-stage difference from OPSD on Qwen3-8B, depend on a favorable seed.

\subsection{Evaluation under the Thinking Mode}
\label{app:nothink-think}

\begin{figure}[h]
\centering
\includegraphics[width=\linewidth]{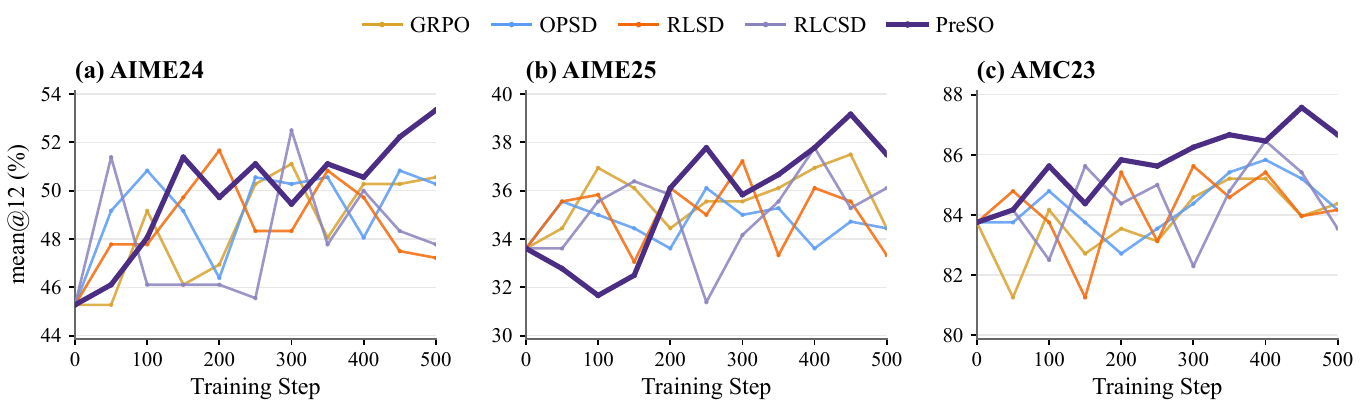}
\vspace{-12pt}
\caption{\textbf{Accuracy over training under the thinking mode.} mean@12 across training steps for Qwen3-1.7B trained in the non-thinking mode and evaluated in the thinking mode.}
\label{fig:train_nothink_eval_think}
\end{figure}

To further assess the robustness of \ours{}, we evaluate the model under the thinking mode. Specifically, with Qwen3-1.7B, both the student and the teacher are run in the non-thinking mode during training, while evaluation is performed in the thinking mode. As shown in Table~\ref{tab:train_nothink_eval_think} and Figure~\ref{fig:train_nothink_eval_think}, \ours{} attains the highest average at the best checkpoint ($60.0$, $+5.8$ over the base) and stays ahead of every baseline, so the gains hold under the thinking mode. The thinking-mode evaluation also sharpens the difference in stability. From the best to the last checkpoint, \ours{} loses only $0.8$ points on average ($60.0 \rightarrow 59.2$), whereas RLCSD drops from $58.9$ to $55.8$ and RLSD from $58.2$ to $54.9$, with several benchmarks ending below the base. \ours{} holds most of its peak to the end, consistent with the stability seen in our main experiments.

\begin{table*}[h]
\caption{\textbf{Accuracy under the thinking mode.} Qwen3-1.7B trained in the non-thinking mode and evaluated in the thinking mode, reporting best- and last-checkpoint mean@12 on AIME24, AIME25, and AMC23. \ours{} achieves the highest average at both checkpoints.}
\centering
\small
\renewcommand{\arraystretch}{1.20}
\setlength{\tabcolsep}{5pt}
\resizebox{\textwidth}{!}{%
\begin{tabular}{l|cccc|cccc}
\toprule
\multirow{2}{*}{\textbf{Method}}
& \multicolumn{4}{c|}{\textbf{Best CKPT}}
& \multicolumn{4}{c}{\textbf{Last CKPT}} \\
\cmidrule(lr){2-5}
\cmidrule(lr){6-9}
& \textbf{AIME24}
& \textbf{AIME25}
& \textbf{AMC23}
& \textbf{Avg.}
& \textbf{AIME24}
& \textbf{AIME25}
& \textbf{AMC23}
& \textbf{Avg.} \\
\midrule

\textbf{Base}
& 45.3 \basex{0.0}
& 33.6 \basex{0.0}
& 83.8 \basex{0.0}
& 54.2 \basex{0.0}
& 45.3 \basex{0.0}
& 33.6 \basex{0.0}
& 83.8 \basex{0.0}
& 54.2 \basex{0.0} \\

\hspace*{0.5em}+ GRPO
& 51.1 \uprise{5.8}
& 37.5 \uprise{3.9}
& 85.2 \uprise{1.4}
& 57.9 \uprise{3.7}
& \underline{50.6} \uprise{5.3}
& 34.4 \uprise{0.8}
& \underline{84.4} \uprise{0.6}
& \underline{56.5} \uprise{2.3} \\

\hspace*{0.5em}+ OPSD
& 50.8 \uprise{5.5}
& 36.1 \uprise{2.5}
& 85.8 \uprise{2.0}
& 57.6 \uprise{3.4}
& 50.3 \uprise{5.0}
& 34.4 \uprise{0.8}
& 84.2 \uprise{0.4}
& 56.3 \uprise{2.1} \\

\hspace*{0.5em}+ RLSD
& 51.7 \uprise{6.4}
& 37.2 \uprise{3.6}
& 85.6 \uprise{1.8}
& 58.2 \uprise{4.0}
& 47.2 \uprise{1.9}
& 33.3 \down{0.3}
& 84.2 \uprise{0.4}
& 54.9 \uprise{0.7} \\

\hspace*{0.5em}+ RLCSD
& \underline{52.5} \uprise{7.2}
& \underline{37.8} \uprise{4.2}
& \underline{86.5} \uprise{2.7}
& \underline{58.9} \uprise{4.7}
& 47.8 \uprise{2.5}
& \underline{36.1} \uprise{2.5}
& 83.5 \down{0.3}
& 55.8 \uprise{1.6} \\

\rowcolor{blue!10}
\hspace*{0.5em}+ \textbf{PreSO}
& \textbf{53.3} \uprise{8.0}
& \textbf{39.2} \uprise{5.6}
& \textbf{87.6} \uprise{3.8}
& \textbf{60.0} \uprise{5.8}
& \textbf{53.3} \uprise{8.0}
& \textbf{37.5} \uprise{3.9}
& \textbf{86.7} \uprise{2.9}
& \textbf{59.2} \uprise{5.0} \\

\bottomrule
\end{tabular}%
}
\label{tab:train_nothink_eval_think}
\end{table*}

\subsection{Out-of-Domain Generalization}
\label{app:ood}

To assess whether math-only training with \ours{} affects unrelated
abilities, we evaluate Qwen3-8B and Qwen3-1.7B trained on
DeepMath-103K on two out-of-domain task families: medicine
(RaR-Medicine~\citep{rar}, HealthBench~\citep{healthbench}, and
LLMEval-Med~\citep{llm_eval_med}) and instruction following
(IFBench~\citep{ifbench} and IFEval~\citep{ifeval}). We score the
open-ended medical responses using Qwen3-32B as a judge and use the
rule-based verifiers provided by the instruction-following benchmarks.
Table~\ref{tb:ood} shows that Qwen3-8B remains close to or slightly
above the base model on all five benchmarks, with differences of at
most $1.2$ points. On Qwen3-1.7B, the changes are mixed and range
from $-1.6$ to $+0.8$ points. Thus, across the evaluated tasks, the
in-domain gains from math training are not accompanied by a
substantial loss in out-of-domain performance.

\begin{table*}[h]
\caption{\textbf{Out-of-domain evaluation of \ours{}.}
Qwen3-8B and Qwen3-1.7B are trained only on DeepMath and evaluated on medical and instruction-following benchmarks (\%).}
\centering
\renewcommand{\arraystretch}{1.25}
\resizebox{\textwidth}{!}{
\begin{tabular}{l|ccc|cc}
\toprule
\multirow{2}{*}{\textbf{Method}}
& \multicolumn{3}{c|}{\textbf{Medicine}}
& \multicolumn{2}{c}{\textbf{Instruction Following}} \\
\cmidrule(lr){2-4}
\cmidrule(lr){5-6}
& \textbf{RaR-Medicine}
& \textbf{HealthBench}
& \textbf{LLMEval-Med}
& \textbf{IFBench}
& \textbf{IFEval} \\
\midrule

\textbf{Qwen3-8B}
& \textbf{56.4}\basex{0.0}
& 62.4\basex{0.0}
& 42.4\basex{0.0}
& 29.0\basex{0.0}
& 86.1\basex{0.0} \\

\hspace*{1em}+ \textbf{PreSO}
& 56.3\down{0.1}
& \textbf{62.5}\uprise{0.1}
& \textbf{43.6}\uprise{1.2}
& \textbf{29.7}\uprise{0.7}
& \textbf{87.3}\uprise{1.2} \\

\midrule

\textbf{Qwen3-1.7B}
& 39.4\basex{0.0}
& 51.1\basex{0.0}
& \textbf{25.5}\basex{0.0}
& \textbf{23.3}\basex{0.0}
& \textbf{73.0}\basex{0.0} \\

\hspace*{1em}+ \textbf{PreSO}
& \textbf{40.2}\uprise{0.8}
& \textbf{51.3}\uprise{0.2}
& 24.0\down{1.5}
& 21.7\down{1.6}
& 72.3\down{0.7} \\

\bottomrule
\end{tabular}
}
\label{tb:ood}
\end{table*}

\subsection{Ablation on token selection}
\ours{} matches representations on the last $512$ tokens of each
response. On Qwen3-1.7B, we compare this choice with the first
$512$ tokens, a uniform sample of $512$ positions, and all tokens
as a higher-budget reference. Table~\ref{tab:token_selection}
shows that the last-token variant achieves the highest final
accuracy on AIME24 and AIME25 and ties the full variant on AMC23.
On AIME24, it changes only slightly from its peak
($18.9\!\to\!18.6$), while the first, uniform, and full variants
decline from $15.6$, $17.2$, and $18.1$ to $11.7$, $13.6$, and
$14.7$, respectively. One possible explanation is that later
positions lie closer to answer formation, where privileged guidance
may be more consequential. The ablation establishes the advantage
of last-token selection in this setting without isolating its
mechanism.

\begin{table*}[h]
\caption{\textbf{Ablation study of token selection in PreSO (Qwen3-1.7B).}
Accuracy (\%) at the best and last checkpoints. Best and second-best results are
marked in \textbf{bold} and \underline{underlined}, respectively.}
\centering
\small
\renewcommand{\arraystretch}{1.15}
\setlength{\tabcolsep}{5pt}
\resizebox{\textwidth}{!}{%
\begin{tabular}{l|cccc|cccc}
\toprule
\multirow{2}{*}{\textbf{Method}}
& \multicolumn{4}{c|}{\textbf{Best CKPT}}
& \multicolumn{4}{c}{\textbf{Last CKPT}} \\
\cmidrule(lr){2-5}
\cmidrule(lr){6-9}
& \textbf{AIME24}
& \textbf{AIME25}
& \textbf{AMC23}
& \textbf{Avg.}
& \textbf{AIME24}
& \textbf{AIME25}
& \textbf{AMC23}
& \textbf{Avg.} \\
\midrule

\textbf{Qwen3-1.7B}
& 12.8 \basex{0.0}
& 9.4 \basex{0.0}
& 44.8 \basex{0.0}
& 22.3 \basex{0.0}
& 12.8 \basex{0.0}
& 9.4 \basex{0.0}
& 44.8 \basex{0.0}
& 22.3 \basex{0.0} \\

\rowcolor{blue!10}
\hspace*{1em}+ \textbf{PreSO (last-layer last-token)}
& \textbf{18.9} \uprise{6.1}
& \textbf{15.6} \uprise{6.2}
& \textbf{54.6} \uprise{9.8}
& \textbf{29.7} \uprise{7.4}
& \textbf{18.6} \uprise{5.8}
& \textbf{15.6} \uprise{6.2}
& \textbf{52.7} \uprise{7.9}
& \textbf{29.0} \uprise{6.7} \\

\hspace*{1em}+ PreSO (last-layer full-token)
& \underline{18.1} \uprise{5.3}
& 13.6 \uprise{4.2}
& \underline{52.7} \uprise{7.9}
& \underline{28.1} \uprise{5.8}
& \underline{14.7} \uprise{1.9}
& \underline{11.7} \uprise{2.3}
& \textbf{52.7} \uprise{7.9}
& \underline{26.4} \uprise{4.1} \\

\hspace*{1em}+ PreSO (last-layer first-token)
& 15.6 \uprise{2.8}
& 13.6 \uprise{4.2}
& 50.6 \uprise{5.8}
& 26.6 \uprise{4.3}
& 11.7 \down{1.1}
& 9.4 \basex{0.0}
& 47.9 \uprise{3.1}
& 23.0 \uprise{0.7} \\

\hspace*{1em}+ PreSO (last-layer uniform-token)
& 17.2 \uprise{4.4}
& \underline{14.4} \uprise{5.0}
& 52.5 \uprise{7.7}
& 28.0 \uprise{5.7}
& 13.6 \uprise{0.8}
& 10.6 \uprise{1.2}
& \underline{50.6} \uprise{5.8}
& 24.9 \uprise{2.6} \\

\bottomrule
\end{tabular}%
}
\label{tab:token_selection}
\end{table*}

\section{More Analysis}
\label{app:more_analysis}
\subsection{Training Efficiency Analysis}
\label{app:efficiency}
As noted in our limitations, \ours{} produces longer rollouts, which
increase its training cost. Table~\ref{tb:efficiency} reports wall-clock
time under the same setup. \ours{} is the slowest method at both scales,
taking $6$ hours $18$ minutes on Qwen3-1.7B and $10$ hours $36$ minutes
on Qwen3-8B, compared with $4$ hours $27$ minutes and $8$ hours
$45$ minutes for GRPO. Longer rollouts increase generation time and
likely contribute substantially to this gap, although these totals
do not isolate generation time from representation-matching overhead.
Reducing training cost while retaining the accuracy gains remains
future work.

\begin{wraptable}{r}{0.4\textwidth}
\centering
\vspace{-\baselineskip}
\begin{tabular}{lcc}
\toprule
Method & 1.7B & 8B \\
\midrule
GRPO  & 4h 27m & 8h 45m \\
OPSD  & 4h 44m & 10h 21m \\
RLSD  & 4h 40m & 8h 50m \\
RLCSD & 5h 09m & 9h 11m \\
\ours{} & 6h 18m & 10h 36m \\
\bottomrule
\end{tabular}
\caption{Training wall-clock time on Qwen3-1.7B and Qwen3-8B.}
\label{tb:efficiency}
\vspace{-\baselineskip}
\end{wraptable}

To assess whether the gains persist under a time budget, we evaluate
the available checkpoints closest to $4$ and $8$ hours
(Table~\ref{tab:equal_training_time}). \ours{} achieves the highest
average at both budgets and both model scales. On Qwen3-8B, it reaches
$40.6$ and $45.5$, compared with $38.7$ and $38.5$ for the strongest
baseline, RLSD. On Qwen3-1.7B, it reaches $26.8$ and $28.6$. The
Qwen3-1.7B runs finish before $8$ hours, so their $8$-hour entries
report final checkpoints. On Qwen3-8B, \ours{} continues to improve
between the two budgets while the baselines are flat or decline.
Thus, its accuracy advantage remains under the evaluated wall-clock
budgets despite its higher cost per training run.

\begin{table*}[h]
\caption{Mathematical reasoning accuracy (\%) under 4-hour and
8-hour training budgets. Each result uses the available checkpoint
closest to the specified budget. For Qwen3-1.7B, training finishes
before 8 hours, so the 8-hour columns use the final checkpoint.}
\centering
\small
\renewcommand{\arraystretch}{1.12}
\setlength{\tabcolsep}{5pt}
\resizebox{\textwidth}{!}{%
\begin{tabular}{l*{12}{c}}
\toprule
\multirow{2}{*}{\textbf{Method}}
& \multicolumn{2}{c}{\textbf{AIME24}}
& \multicolumn{2}{c}{\textbf{AIME25}}
& \multicolumn{2}{c}{\textbf{AIMO}}
& \multicolumn{2}{c}{\textbf{HMMT25}}
& \multicolumn{2}{c}{\textbf{AMC23}}
& \multicolumn{2}{c}{\textbf{Avg.}} \\
\cmidrule(lr){2-3}
\cmidrule(lr){4-5}
\cmidrule(lr){6-7}
\cmidrule(lr){8-9}
\cmidrule(lr){10-11}
\cmidrule(lr){12-13}
& \textbf{4h} & \textbf{8h}
& \textbf{4h} & \textbf{8h}
& \textbf{4h} & \textbf{8h}
& \textbf{4h} & \textbf{8h}
& \textbf{4h} & \textbf{8h}
& \textbf{4h} & \textbf{8h} \\
\midrule

\textbf{Qwen3-8B}
& \multicolumn{2}{c}{25.6}
& \multicolumn{2}{c}{18.6}
& \multicolumn{2}{c}{61.4}
& \multicolumn{2}{c}{10.8}
& \multicolumn{2}{c}{68.3}
& \multicolumn{2}{c}{36.9} \\

\hspace*{1em}+ GRPO
& 27.2 & 27.5
& 18.9 & 18.3
& \underline{63.0} & 61.7
& 10.6 & 10.6
& 67.7 & \underline{69.2}
& 37.5 & 37.5 \\

\hspace*{1em}+ OPSD
& 25.0 & 17.2
& 18.3 & 12.8
& 58.5 & 43.2
& 10.3 & 3.9
& 65.6 & 54.2
& 35.5 & 26.3 \\

\hspace*{1em}+ RLSD
& \underline{30.0} & \underline{30.0}
& \underline{19.4} & 20.3
& 61.8 & \underline{63.0}
& \underline{11.7} & \underline{11.7}
& 70.6 & 67.5
& \underline{38.7} & \underline{38.5} \\

\hspace*{1em}+ RLCSD
& 26.7 & 28.1
& \underline{19.4} & \underline{22.2}
& 62.0 & 61.0
& 8.6 & 9.4
& \underline{72.7} & 68.5
& 37.9 & 37.8 \\

\hspace*{1em}+ \textbf{PreSO}
& \textbf{33.3} & \textbf{37.0}
& \textbf{19.7} & \textbf{26.7}
& \textbf{63.6} & \textbf{70.4}
& \textbf{12.8} & \textbf{16.4}
& \textbf{73.5} & \textbf{77.1}
& \textbf{40.6} & \textbf{45.5} \\

\midrule

\textbf{Qwen3-1.7B}
& \multicolumn{2}{c}{12.8}
& \multicolumn{2}{c}{9.4}
& \multicolumn{2}{c}{40.3}
& \multicolumn{2}{c}{5.6}
& \multicolumn{2}{c}{44.8}
& \multicolumn{2}{c}{22.6} \\

\hspace*{1em}+ GRPO
& 12.8 & 13.1
& 10.3 & \underline{11.1}
& 39.1 & 41.5
& \underline{5.6} & 6.1
& 43.5 & 46.7
& 22.3 & 23.7 \\

\hspace*{1em}+ OPSD
& \underline{14.4} & 13.9
& 10.0 & \underline{11.1}
& 40.4 & \underline{42.3}
& 4.7 & 5.0
& 46.0 & \underline{47.7}
& 23.1 & 24.0 \\

\hspace*{1em}+ RLSD
& 14.2 & \underline{16.4}
& \underline{10.8} & 8.9
& \underline{41.7} & 41.2
& \underline{5.6} & \underline{6.4}
& \underline{47.9} & 47.5
& \underline{24.0} & \underline{24.1} \\

\hspace*{1em}+ RLCSD
& 11.4 & 13.6
& 10.3 & 10.6
& 41.3 & 42.0
& 5.0 & 6.1
& 44.6 & 45.8
& 22.5 & 23.6 \\

\hspace*{1em}+ \textbf{PreSO}
& \textbf{18.3} & \textbf{18.6}
& \textbf{13.6} & \textbf{15.6}
& \textbf{45.2} & \textbf{48.2}
& \textbf{6.1} & \textbf{8.1}
& \textbf{50.6} & \textbf{52.7}
& \textbf{26.8} & \textbf{28.6} \\

\bottomrule
\end{tabular}%
}
\label{tab:equal_training_time}
\end{table*}

\subsection{Abstraction versus Information Content}
\label{app:concept}

\begin{figure}[h]
\centering
\includegraphics[width=\linewidth]{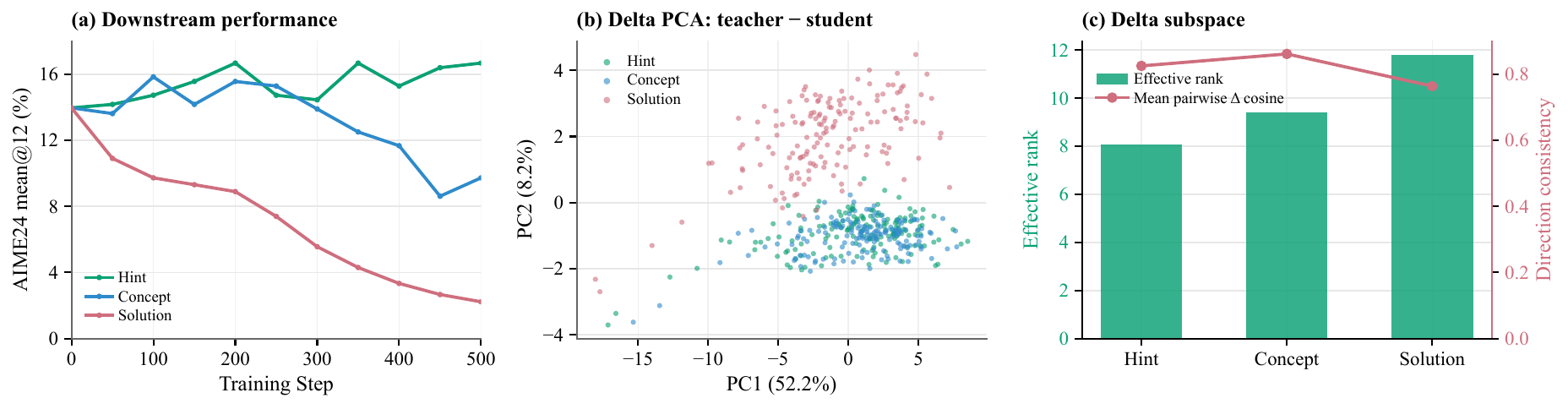}
\vspace{-12pt}
\caption{\textbf{Abstraction versus information content of privileged sources.}
Comparison of the hint, concept, and solution sources on Qwen3-1.7B.
(a) Downstream mean@12 on AIME24 over training.
(b) Teacher$-$student shifts projected onto the PCA axes already fitted in
Fig.~\ref{fig:rep_pi_analysis}(b); axis labels report the variance those axes explain on
the original hint/solution shifts and are reproduced here for reference.
(c) Effective rank and direction consistency of each shift subspace.
The concept source gives a low-rank, directionally consistent shift yet trails
the hint downstream, showing that abstraction is not monotone in benefit.}
\label{fig:concept}
\end{figure}

The main text compares compact hints with full solutions, finding that
solutions induce more dispersed shifts and less stable downstream
performance. This raises the opposite question of whether making
privileged information even more abstract would help. We therefore
introduce a \emph{concept} source that names the broad mathematical
topic without providing a solving strategy or answer, and compare all
three sources on Qwen3-1.7B (Fig.~\ref{fig:concept}). On AIME24, the
concept source underperforms hints throughout training and declines at
later checkpoints. Yet its shift has low effective rank and directional
consistency comparable to the hint-induced shift
(Fig.~\ref{fig:concept}c). Concentration and consistency alone therefore
do not guarantee useful guidance. One possible explanation is that a
concept omits actionable strategy-level information despite inducing
a coherent shift. Among these three sources, hints offer the best
observed balance between limiting solution-specific detail and
providing guidance the student can use.

\subsection{Stability of the Warm-up Subspace}
\label{app:subspace-stability}

\ours{} estimates the privileged subspace during warm-up and freezes
it thereafter. We ask whether it continues to capture the
privileged shift as training proceeds. At each checkpoint $s$, we
compute $\Delta_{i,t}^{(s)}
=\bar h_{i,t}^{\mathrm{PI},(s)}
-\bar h_{i,t}^{\mathrm{noPI},(s)}$
on held-out problem $i$ and rollout position $t$, where both
representations are $\ell_2$-normalized. For a candidate set of
cluster-specific projectors $P$, we measure the fraction of shift
energy retained by the projector routed to each problem:
\begin{equation}
R_P(s)=
\frac{\mathbb{E}\lVert P_{c_i}\Delta_{i,t}^{(s)}\rVert_2^2}
{\mathbb{E}\lVert\Delta_{i,t}^{(s)}\rVert_2^2},
\end{equation}
where the expectation is over held-out problems, rollouts, and valid
token positions.

\begin{figure}[!tbp]
    \centering
    \includegraphics[width=0.72\linewidth]{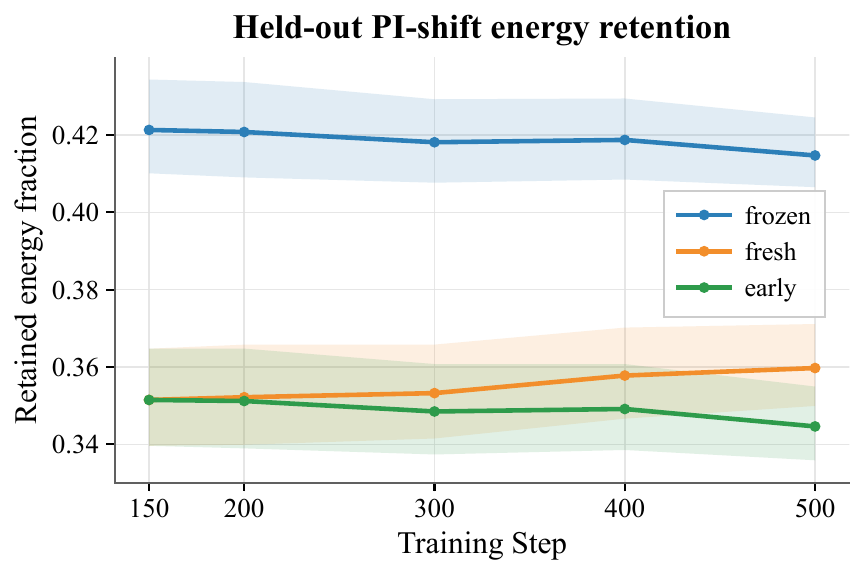}
    \caption{\textbf{Retention of held-out privileged-shift energy.}
    The frozen warm-up subspace retains a nearly constant fraction
    during training on Qwen3-1.7B.}
    \label{fig:subspace-stability}
\end{figure}

We compare three same-rank subspaces on identical held-out shifts.
The \texttt{frozen} subspace is estimated from the 1200-problem
warm-up calibration set and used during training. The \texttt{fresh}
reference is re-estimated at each checkpoint, while the
\texttt{early} reference is estimated at an early checkpoint and
held fixed. Both references use a cluster-stratified diagnostic
split of 62 problems. All three are evaluated on a disjoint split
of 65 problems, so the evaluated shifts are not used to estimate
the subspaces. Figure~\ref{fig:subspace-stability} reports results
on Qwen3-1.7B with cluster-stratified bootstrap $95\%$ intervals
over the evaluation problems.

The frozen subspace retains a nearly constant fraction of held-out
shift energy, changing from about $0.42$ to $0.41$ across training.
The relevant observation is this stability, not the absolute
fraction retained. At a fixed rank, retention need not approach
one, and this measure does not identify the retained energy as
useful or the discarded energy as noise. The \texttt{fresh} and
\texttt{early} references retain less energy, but they are fitted
on 62 rather than 1200 problems. Their gap from \texttt{frozen}
therefore cannot establish that freezing outperforms periodic
re-estimation.

These results support using a frozen warm-up subspace over the
observed training run because its coverage of held-out shifts does
not appreciably decline. They do not establish that its retained
directions are all task-relevant or that freezing causes
\ours{}'s downstream stability. A direct comparison with
periodic re-estimation would require matched fitting sets.
The diagnostic and calibration problems come from the
same corpus, so this experiment assesses within-corpus stability.

\subsection{Visualization of Shift Clusters and Representative Problems}
\label{app:cluster-vis}

\begin{figure}[!tbp]
    \centering
    \includegraphics[width=0.62\linewidth]{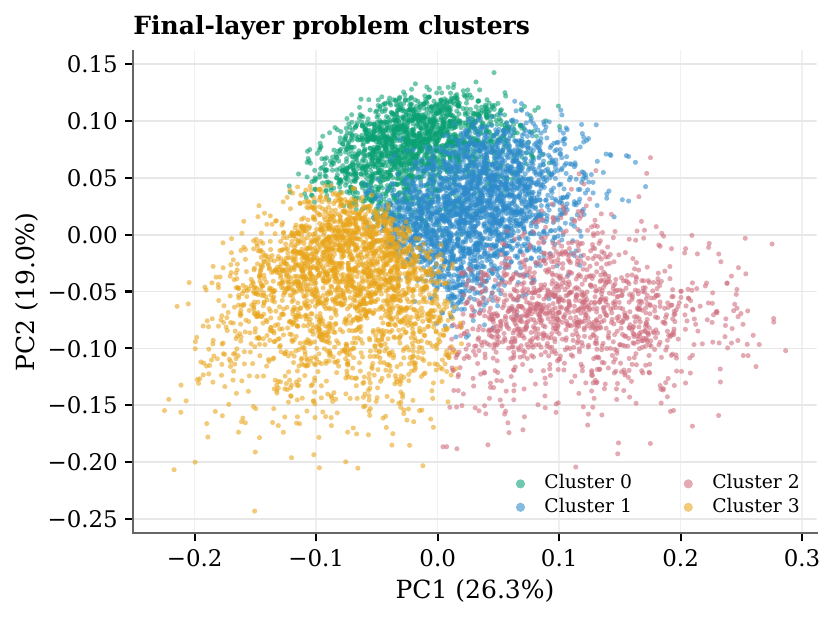}
    \caption{\textbf{PCA visualization of privileged shifts on
    Qwen3-1.7B.} Colors indicate $k$-means cluster assignments
    obtained from the shifts in representation space.}
    \label{fig:cluster-vis}
\end{figure}

To visualize how privileged shifts vary across problems, we compute
a shift for each held-out DeepMath problem on Qwen3-1.7B. We cluster
these shifts with $k$-means in representation space and project them
onto two principal components for display in
Figure~\ref{fig:cluster-vis}. Each point represents a problem, and
its color indicates its cluster assignment. The projection offers
a qualitative view of how the shifts are organized. Since the
clusters are fitted on the same shifts, their visible separation
is not an independent test of cross-problem recurrence.

Table~\ref{tab:cluster-cases} lists descriptive keywords and one
representative problem for each cluster. Some clusters are associated
with broad mathematical themes, such as algebraic manipulation or
counting. These examples are consistent with content-dependent
organization in the shift space. They provide intuition for comparing
related problems, but do not establish that a cluster's leading
directions isolate useful knowledge or have a unique semantic
interpretation.

\begin{table*}[h]
    \centering
    \caption{\textbf{Cluster examples on DeepMath with Qwen3-1.7B.}
For each cluster we report its size and share of held-out problems, the keywords most associated with it, and one example problem.}
    \label{tab:cluster-cases}

    \small
    \setlength{\tabcolsep}{5pt}
    \renewcommand{\arraystretch}{1.25}

    \begin{tabularx}{\textwidth}{
        @{}
        >{\raggedright\arraybackslash}p{0.10\textwidth}
        >{\raggedright\arraybackslash}p{0.17\textwidth}
        >{\raggedright\arraybackslash}p{0.18\textwidth}
        >{\raggedright\arraybackslash}X
        @{}
    }
        \toprule
        \textbf{Cluster}
        & \textbf{Descriptive label}
        & \textbf{Representative keywords}
        & \textbf{Example problem} \\
        \midrule

        \textbf{C0}\newline
        $n=6{,}470$\newline
        (20.92\%)
        &
        Constrained problem solving and optimization
        &
        \textit{value, equation, functions, integer, real numbers}
        &
        Let $f:\mathbb{R}\to\mathbb{R}$ be three times
        differentiable, with at least five distinct zeros.
        Determine the minimum number of zeros of
        $f+6f'+12f''+8f'''$.
        \\

        \addlinespace[5pt]

        \textbf{C1}\newline
        $n=8{,}424$\newline
        (27.24\%)
        &
        Statement validity and existence
        &
        \textit{determine, exists, true, false,
        statement, justification}
        &
        Let $B_1,B_2\subset\mathbb{R}^3$ be topological
        balls with nonempty interiors. Determine whether
        the connected components of $B_1\cap B_2$ are
        also topological balls.
        \\

        \addlinespace[5pt]

        \textbf{C2}\newline
        $n=9{,}945$\newline
        (32.16\%)
        &
        Integrals, limits, and series
        &
        \textit{evaluate, limit, integral, residue,
        series, contour}
        &
        Evaluate the complex integral
        $\int_{\gamma}\frac{z}{\overline{z}}\,dz$,
        where $\gamma$ is the boundary of the region
        $\{z\in\mathbb{C}:1<|z|<2,\,
        \operatorname{Im}z>0\}$.
        \\

        \addlinespace[5pt]

        \textbf{C3}\newline
        $n=6{,}083$\newline
        (19.67\%)
        &
        Implications and necessary properties
        &
        \textit{does, imply, necessarily, space,
        compact, finite}
        &
        Let $s>t$ and let $V$ be a linear subspace of
        $H^s(\mathbb{R}^n)$. If there exists a constant
        $C$ such that
        $\|f\|_{H^s}\leq C\|f\|_{H^t}$
        for every $f\in V$, must $V$ be finite-dimensional?
        \\

        \bottomrule
    \end{tabularx}
\end{table*}

\subsection{Case Study}
\label{app:case-study}
To make the effect of \ours{} concrete, we compare the base model with \ours{} on two problems aligned with the themes of clusters in Section~\ref{app:cluster-vis}, with condensed excerpts in Figure~\ref{fig:case-study}.

The first case comes from cluster C0, which groups constrained optimization problems. The task is to maximize the real part of an expression in $z$ under $|z|=4$. The base model makes a sign error early and, without noticing, optimizes the wrong objective to a wrong value. \ours{} keeps the intended objective, parameterizes $z$ on the circle $|z|=4$, and reduces the problem to an amplitude computation with the correct answer. The improvement here comes from carrying a coherent line of reasoning all the way through, the solution structure that recurs across C0.

The second case comes from cluster C1, which groups validity and existence problems, and asks for all real $k$ under which a system of distance constraints has exactly one solution. The base model places the real parameter on the wrong axis and carries this inconsistent geometry forward to a wrong answer. \ours{} starts similarly, but partway through detects that one distance is incompatible with its current picture, rebuilds the geometry from the squared distances, and applies the tangency condition to reach the correct answer. Beyond following the right structure, this case shows the model checking its own intermediate reasoning and revising it when it does not hold together.

Together the two cases suggest that \ours{} does not merely memorize answers but internalizes the reasoning behavior of each cluster, both sustaining a correct line of reasoning and catching and correcting its own missteps. We view these examples as illustrative rather than a systematic evaluation of reasoning quality.

\BasePreSOCombinedCases

\section{Experiment details}
\subsection{Implementation details}
\label{app:training-details}

We conduct all experiments with the verl~\citep{verl} training framework. All methods share the same optimizer, batch size, rollout budget, and sequence lengths, so that differences in performance come from the objective rather than the training budget. At evaluation we decode $12$ samples per problem with nucleus sampling and score them with a rule-based verifier, reporting mean@12. The full configuration is listed in Table~\ref{tb:hparams}.

\begin{table*}[h]
\centering
\caption{Full hyperparameter configuration for \ours{} and all baselines.}
\label{tb:hparams}
\begin{tabularx}{\textwidth}{@{}l >{\raggedright\arraybackslash}X@{}}
\toprule
\textbf{Settings} & \textbf{Hyperparameters} \\
\midrule
\textbf{General}
& Learning rate $=1\times10^{-6}$, Train batch size $=8$, Rollouts per prompt $G=8$ \\
& Sampling temperature $=1.0$, Top-$p=1.0$, Total steps $=500$ \\
& Max prompt length $=2{,}048$, Max response length $=16{,}384$ \\
\midrule
\textbf{Evaluation}
& Samples per problem $=12$, Temperature $=0.7$, Top-$p=0.95$ \\
& Max prompt length $=2{,}048$, Max generation length $=38{,}912$ \\
\midrule
\textbf{GRPO}
& PPO clip $\epsilon=0.2$, Reference-policy KL coefficient $\lambda_{\text{KL}}=0.001$ \\
\midrule
\textbf{OPSD}
& JSD mixing coefficient $\beta_{\text{mix}}=0$ (forward KL), Distillation top-$k=100$ \\
& KL contribution clip $=0.05$, Tail aggregation enabled \\
\midrule
\textbf{RLSD}
& PPO clip $\epsilon=0.2$, Evidence-weight clip $\epsilon_w=0.2$ \\
& Initial evidence mixing $\lambda_0=0.5$, Mixing decay steps $=50$ \\
\midrule
\textbf{RLCSD}
& PPO clip $\epsilon=0.2$, Evidence temperature $\tau=0.02$, Residual scale $\beta_{\text{res}}=1.0$ \\
& Residual mixing $\lambda=0.5$, Selection threshold $\delta=0.02$, Selected-token loss weight $\eta=1.0$, Negative hints per sample $K_{\max}=4$, Residual clip $[-2,2]$ \\
\midrule
\textbf{\ours{}}
& Subspace rank $r=128$, Content routes $K=4$, Warm-up steps $W=150$, Loss coefficient $\lambda=0.1$, Distillation layer: final, Supervised positions: last $512$ \\
\bottomrule
\end{tabularx}
\end{table*}

\subsection{Prompt Details}
\label{app:prompts}

For completeness we list the prompts used in our experiment. The first is the template that conditions the self-teacher on the privileged information to produce the privileged representation, and the second and third are the templates used to annotate each training problem with a hint and with a concept respectively. Table~\ref{tab:hint-concept-cases} shows several examples contrasting the full solution, the hint, and the concept for the same problem.

\TeacherModelPrompt
\HintGenerationPrompt
\ConceptGenerationPrompt

\begin{table*}[t]
\centering
\caption{Examples of math problems, solutions, hints, and concepts.}
\label{tab:hint-concept-cases}

\small
\renewcommand{\arraystretch}{1.15}
\setlength{\tabcolsep}{6pt}
\setlength{\abovedisplayskip}{4pt}
\setlength{\belowdisplayskip}{4pt}

\begin{tabularx}{\textwidth}{
  @{}
  >{\bfseries\raggedright\arraybackslash}p{0.12\textwidth}
  >{\raggedright\arraybackslash}X
  @{}
}
\toprule

\multicolumn{2}{@{}l}{
  \textbf{Case 1: Symmetric polynomials}
}
\\
\midrule

Problem &
Given that
\[
\begin{aligned}
\alpha+\beta+\gamma &= 14,\\
\alpha^2+\beta^2+\gamma^2 &= 84,\\
\alpha^3+\beta^3+\gamma^3 &= 584,
\end{aligned}
\]
find \(\max\{\alpha,\beta,\gamma\}\).
\\
\midrule

Solution &
Let \(s_2=\alpha\beta+\beta\gamma+\gamma\alpha\)
and \(s_3=\alpha\beta\gamma\).
The identities for the sums of squares and cubes give
\[
\begin{aligned}
84 &= 14^2-2s_2,
& s_2 &= 56,\\
584 &= 14^3-3\cdot14\cdot56+3s_3,
& s_3 &= 64.
\end{aligned}
\]
Thus, \(\alpha,\beta,\gamma\) are the roots of
\[
x^3-14x^2+56x-64=(x-2)(x-4)(x-8).
\]
Therefore, \(\max\{\alpha,\beta,\gamma\}=\boxed{8}\).
\\
\midrule

Hint &
Use the power sums to recover the elementary symmetric sums
via standard identities, then view \(\alpha,\beta,\gamma\)
as roots of a cubic determined by those symmetric sums.
Once that polynomial is formed, its largest root is the
quantity you want.
\\
\midrule

Concept &
Symmetric polynomials and polynomial roots.
\\

\midrule[\heavyrulewidth]

\multicolumn{2}{@{}l}{
  \textbf{Case 2: Trigonometric simplification}
}
\\
\midrule

Problem &
Simplify
\[
(\tan\alpha+\cot\alpha)^2
-
(\tan\alpha-\cot\alpha)^2.
\]
\\
\midrule

Solution &
By the difference of squares formula,
\[
\begin{aligned}
&(\tan\alpha+\cot\alpha)^2
-(\tan\alpha-\cot\alpha)^2\\
&\qquad=(2\cot\alpha)(2\tan\alpha)\\
&\qquad=4\cot\alpha\tan\alpha
=\boxed{4},
\end{aligned}
\]
using the reciprocal identity
\(\cot\alpha\tan\alpha=1\).
\\
\midrule

Hint &
Recognize this as a difference of squares, so think about
factoring the entire expression instead of expanding both
binomials. Then use the reciprocal relationship between
tangent and cotangent to simplify the product.
\\
\midrule

Concept &
Trigonometric identities and algebraic simplification.
\\
\bottomrule

\end{tabularx}
\end{table*}

\end{document}